%% file: main.tex
\documentclass{article}

\usepackage{amsmath,amsfonts}
\usepackage{algorithmic}
\usepackage{algorithm}
\usepackage{array}
\usepackage{textcomp}
\usepackage{stfloats}
\usepackage{url}
\usepackage{verbatim}
\usepackage{cite}
\usepackage{svg}
\usepackage{multirow}
\usepackage{xcolor}
\usepackage{soul}

\usepackage{jfrExamplee}
\usepackage{graphicx}
\usepackage{apalike}
\usepackage{setspace}

\usepackage{comment}
\usepackage{float,longtable}
\usepackage{lipsum}
\usepackage{pdflscape}
\usepackage{tabularx}
\usepackage{adjustbox}
\usepackage{subcaption}
\usepackage{rotating}
\usepackage{longtable}
\usepackage{xcolor}

\usepackage{lscape}

\usepackage{makecell}
\usepackage[plainpages=false,hypertexnames=true,pdfnewwindow=true,colorlinks=true,citecolor=black,linkcolor=black,urlcolor=blue,filecolor=blue]{hyperref}

\usepackage{amssymb}
\usepackage{pifont}

\title{Autonomous Strawberry Picking Robotic System}

\author{
Soran Parsa\thanks{Soran and Amir equally contributed to this work}\\ 
Lincoln Institute for Agri-food Technology\\ 
University of Lincoln\\
Lincoln, UK\\
\texttt{soran.parsa@gmail.com} \\
\And
Bappaditya Debnath \\
Kings College London \\
London, UK \\
\texttt{b.debnath2017@gmail.com}\\
\AND
 Muhammad Arshad Khan\\ 
Lincoln Institute for Agri-food Technology\\ 
University of Lincoln\\
Lincoln, UK\\
\texttt{MuKhan@lincoln.ac.uk} \\
\And
Amir Ghalamzan E.$^*$\thanks{ Corresponding author}; \thanks{This work was supported by CERES Agri Tech. It is also partially supported by the Centre for Doctoral Training, United Kingdom (CDT) in Agri-Food Robotics (AgriFoRwArdS) Grant reference: EP/S023917/1; Lincoln Agri-Robotics (LAR) funded by Research England.}\\
Lincoln Institute for Agri-food Technology\\ 
University of Lincoln\\
Lincoln, UK\\
{\texttt aghalamzanesfahani@lincoln.ac.uk}\\
}

\begin{document}

\maketitle

\begin{abstract}
Challenges in strawberry picking made selective harvesting robotic technology very demanding. However, the selective harvesting of strawberries is a complicated robotic task forming a few scientific research questions. 
Most available solutions only deal with a specific picking scenario, e.g., picking only a single variety of fruit in isolation. Nonetheless, most economically viable (e.g. high-yielding and/or disease-resistant) varieties of strawberry are grown in dense clusters. The current perception technology in such use cases is inefficient. In this work, we developed a novel system capable of harvesting strawberries with several unique features. These features allow the system to deal with very complex picking scenarios, e.g. dense clusters. Our concept of a modular system makes our system reconfigurable to adapt to different picking scenarios. 
We designed, manufactured, and tested a patented picking head with 2.5 degrees of freedom (two independent mechanisms and one dependent cutting system) capable of removing possible occlusions and harvesting the targeted strawberry without any contact with the fruit flesh to avoid damage and bruising. In addition, we developed a novel perception system to localise strawberries and detect their key points, picking points, and determine their ripeness. For this purpose, we introduced two new datasets. Finally, we tested the system in a commercial strawberry growing field and our research farm with three different strawberry varieties. The results show the effectiveness and reliability of the proposed system. The designed picking head was able to remove occlusions and harvest strawberries effectively. The perception system was able to detect and determine the ripeness of strawberries with 95\% accuracy. In total, the system was able to harvest 87\% of all detected strawberries with a success rate of 83\% for all pluckable fruits. We also discuss a series of open research questions in the discussion section.       
\end{abstract}

\textbf{keywords:} Selective Harvesting; Robotic manipulation; Computer Vision; Motion planning; Precision farming; Agricultural robotics 

\section{Introduction}

Selective harvesting of crops using robotic technology aims to address the societal and economical challenges of agricultural labour shortages. The industry is yet far from an efficient and practical solution. Many aspects of crop harvesting still remain unsolved (scientific and technological) problems. 
The dexterity and efficiency of robotic end-effectors are open questions. Most of the available picking heads (i.e. end-effectors) for selective harvesting are capable of performing only two actions: opening the picking head, and closing the picking head.

Strawberry is a highly valued crop. While the annual retail value of the strawberries industry is over \textdollar 17 Billion globally, producers have to spend over \textdollar 1 Billion for picking (selective harvesting)  only~\cite{pickingcost21}. Factors such as labour shortage, increasing labour costs, and the COVID-19 pandemic are having a negative impact on selective harvesting costs. Therefore, robot-based automated selective harvesting technologies are highly in demand~\cite{duckett2018agricultural}. Over the past decade, both private and public entities have extensively invested to develop commercially viable robotic harvesting technology. Despite the recent investments and funding~\cite{bachus21} and~\cite{sweeper2021} for harvesting high-value crops, many problems are still unsolved which form very interesting scientific questions. Nevertheless, there is not yet a commercially viable robotic technology available for the selective harvesting of strawberries. 

One of the challenges of a desired robotic solution is the picking head they use. While human pickers use the sense of touch and active perception, multiple fingers and two arms for picking strawberries, using a picking head with a single degree of freedom (DOF) does not look sufficient. The available picking heads have limited ability to harvest strawberries in a dense cluster where a ripe strawberry to be picked is occluded. The problem is manifold: (1) a ripe strawberry may not be detected, (2) its segment, ripeness assessment, and location may not be precise, (3) existing picking heads may not be able to reach a ripe strawberry surrounded by other unripe strawberries. We present a robotic picking system capable of addressing these issues

 Moreover, the asymmetric and irregular nature of the stems coming out of the fruit makes it difficult to localise the picking point. Commercially available depth sensors are designed for large objects under controlled lighting conditions. Insufficient quality of depth-sensing technologies makes strawberry picking point localisation on stem intractable. This is especially true under bright sunlight in farm conditions where the depth accuracy decreases further. In addition, the depth sensors are designed to work optimally for distances larger than 50 [cm], and their precision drops to 0 for distances below 15 [cm]. However, for picking point localisation we require precise depth-sensing below the 15 [cm] range. 
 
 This makes the robot perception challenging as some target fruits may be occluded by non-target fruits and leaves. Commercially available depth sensors, e.g, Realsense \emph{D435i}, also make the perception challenging as they are designed for large objects' 3-D perception and controlled lighting conditions. For small fruits under outdoor lighting, the depth maps are not precise. Detecting, segmenting, and localising a ripe fruit to be picked in a complex cluster geometry, under outdoor lighting conditions make strawberry perception a very challenging problem.

 In this paper, we present a robotic system for automated selective harvesting of strawberries which aims to address some of the challenges preventing large-scale commercial deployment of these systems. (A video of the system can be seen in \href{https://www.youtube.com/watch?v=JF4WR6Li-v4&ab_channel=UniversityofLincoln}{this link}.) We designed, prototyped and field-tested our robotic system benefiting from a novel picking head for robotic selective harvesting. The picking head demonstrated the ability to navigate through the cluster to reach a targeted fruit and harvest it successfully. One feature of the picking head that makes it different from the existing technologies, is that it is able to grasp, detach and handle the fruit from its stems without contact with the fruit body. This is important in harvesting and handling soft fruits e.g. strawberries, to reduce bruising and damage and increase fruit shelf time. These characteristics are able to contribute to reducing food waste significantly. 
 
 Our state-of-the-art perception system proved to be effective in detecting and localising fruit in different environments and lighting conditions. Moreover, the localising of the picking point on the stem is challenging. This is due to the nature of the 3D perception devices that work poorly on small objects, or under sunlight conditions. To overcome this challenge we propose a novel Gaussian Process Regression method for picking point error estimation.

 We propose a modular and configurable approach to developing and integrating the robotic system for selective fruit harvesting. The system was reconfigured based on two different harvesting conditions and tested. Unlike other approaches the robotic harvesting system is designed based on a specific requirement and only work in a specific condition, this proposed system is modular and configurable based on the different varieties and growing condition. The remainder of this paper is organised as follows. Section \ref{sec:literature} presents a thorough literature review of the current works. In section \ref{sec:consept} and \ref{sec:endeffector} the system architecture and end-effector design are discussed respectively. Section \ref{sec:perception} presents the perception system. Finally, the field experiments and results are presented in section \ref{sec:results} and are discussed in section \ref{sec:descussion}.

\section{Related Works}
\label{sec:literature}
\subsection{Harvesting systems and manipulators}

Robotic harvesting systems in general are mechanisms that are designed to interact with agricultural crops. A typical robotic harvesting system is equipped with a manipulator usually in form of a serial robotic arm, a custom-designed end-effector for grasping and/or picking the targeted crop, a perception system for detection, and a platform to mount all these sub-components which itself could be an autonomous or semi-autonomous powered mobile platform \cite{arad2020development}.

Off-the-shelf robotic arms and manipulators have proven to be functional and reliable and have been employed largely to develop robotic selective harvesting systems. However, custom-designed manipulators have been emphasised greater than using off-the-shelf ones. Among the off-the-shelf manipulators, six degrees of freedom (DOF) were widely used. It has been studied that additional degrees of freedom were added to or removed from the available controllable DOFs in these off-the-shelf manipulators according to the harvesting requirements. Xiong et.al.\cite{xiong2019development} used a 6-DOF off-the-shelf manipulator for harvesting strawberries where 1-DOF was kept fixed during operation to meet selective harvesting orientation requirement. 

In addition to the single-arm manipulator, multiple-arm robots were also utilised for harvesting scenarios to tackle the complexity of selective harvesting. In particular, dual-arm manipulators were developed to work either collaboratively or as standalone units. Sepulveda et.al.\cite{sepulveda2020robotic} used a dual robotic arm to cooperatively harvest eggplants with very promising and successful results in dealing with occluded fruit conditions. Zhao et.al.\cite{zhao2016dual} also used a dual-arm robot for tomato harvesting which also operated collaboratively. In this proposed configuration, one arm detaches a tomato from its stem while the other arm grips it. Davidson J et.al.\cite{davidson2017dual} utilised a dual-arm mechanism for an apple harvesting robot in which a six-degree of freedom (DOF) apple picker was assisted by a two-DOF catching mechanism. The second arm catches the picked apple and transfers it to storage. This reduces the harvesting cycle time. A few other dual-arm configurations were also presented in~\cite{armada2005prototype,ceres1998design,xiong2020autonomous} to speed up the harvesting cycle. The autonomous kiwi harvesting robot\cite{scarfe2009development} uses four robots in parallel.

The harvesting robot must reach the varying height, widths, and depths of the targeted crop with respect to the base of the manipulator. Hence, the harvesting platform needs a moving base to increase the limited reachable workspace of a robotic manipulator. In addition to a mobile base that can navigate across fields, manipulators were also mounted on the vertical slide(s)~\cite{zhao2016dual,lehnert2017autonomous}\cite{bac2017performance,ling2019dual,baeten2008autonomous}, horizontal slide(s)~\cite{davidson2017dual,silwal2017design,van2002autonomous}, slanting slide~\cite{armada2005prototype}, or on a scissor lift mechanism~\cite{arad2020development}\cite{feng2018design} to enhance the reachability of a robotic arm. Moreover, a forklift vehicle was utilised to enable the cutter mechanism to reach the various height and depths to harvest oranges in orchards~\cite{lee2006development}. 

In addition to rigid mechanisms, other mechanisms were also employed for harvesting~\cite{chowdhary2019soft}. For instance, \cite{tiefeng2015fruit} proposed an elephant trunk-inspired mechanism to harvest fruits. Combined soft and rigid mechanisms have been also tested for agroforestry activities which include harvesting as well.~\cite{chowdhary2019soft}. 

Motion planning and motion control are also important components of a successful selective harvesting robotic system. 
Mghames et al.~\cite{mghames2020interactive} proposed an interactive motion plan to push occluding strawberries away in a cluster to reach a target fruit. Pushing actions are encoded by movement primitives and hand-designed features of pushable obstacles. A set of pushing demonstrations is used to train the motion primitives. Learning from demonstrations~\cite{ragaglia2018robot}, and imitation learning~\cite{osa2018algorithmic} are used in many other contexts. However, their hand-designed features (such as the position of the target and occluding strawberries) may limit the generalisation of such a method. Sanni et al.~\cite{sanni2022deep} proposed deep-Movement Primitives (d-MP) that do not need any hand-designed features and directly map visual information into robot movements based on the observed demonstrations. 
Tafuro et al.~\cite{tafuro2022dpmp} extended d-MP into deep-Probabilistic movement primitives (d-ProMP) in which the model generates a distribution of trajectories given a single image of the scene. 

To control pushing motions occluded camera views are not sufficient and tactile sensings are necessary. Mandil et al.~\cite{mandil2022action} proposed a data-driven tactile predictive model which is then used in~\cite{nazari2022proactive} to proactively control manipulation movements to avoid slip. This framework can be adopted to control pushing actions.

\subsection{End-effectors}

An end-effector is a tool attached to the wrist of a robotic manipulator to harvest the fruit, either by
grasping or gripping the fruit or its peduncle (attachment), detaching it from the parent plant, and eventually
delivering it to the storage. These end-effectors execute the individual actions either simultaneously (eg:
gripping and detaching) or sequentially (gripping/grasping followed by detaching) to perform a successful
harvesting operation. The end-effector of a selective harvesting robot is the unit that directly and physically interacts with crops. Across different end-effector technologies for fruit harvesting, the physical interactions include (i) gripping/grasping the fruit by its peduncle or fruit body (attachment), (ii) detaching from the plant by pulling, twisting, or cutting the peduncle, (iii) facilitating the fruit transport from the detachment location to the storage, and (iv) pushing/parting of fruit in the cluster during detaching action\cite{xiong2020obstacle} that is a recently studied functionality.

Among attachment and detachment actions, some end-effectors perform simultaneous gripping and detachment of the strawberry peduncle using a parallel jaw mechanism~\cite{hayashi2010evaluation,hayashi2014field}. In such a mechanism, one jaw will be shaped in the form of a cutting blade or is provided with a provision to attach detaching blades. One such end-effector design makes use of a suction cup to provide an additional grip by sucking the fruit body to avoid any positional errors during this simultaneous gripping and cutting actions~\cite{hayashi2010evaluation}. Another suction-based approach is used by an end-effector which uses a suction head to grip the fruit body and then rotates so that a blade is positioned on the curved opening of the suction head to trim the peduncle\cite{arima2004strawberry}. Instead of using any cutting blades for trimming the peduncle, the end-effector in \cite{yamamoto2014development} uses a bending action to detach the strawberry after gripping the fruit body with a suction head and two-jaw gripper.

In addition, a thermal-based cutting is reported to be used by another end-effector that uses an electrically heated wire on the gripping jaw to cut the peduncle~\cite{feng2012new}. In this end-effector, once the fruit body is gripped by the suction cup to position the peduncle between the two jaws (cutting device), the jaws then close and trim the peduncle using the heated wire. The end effector developed by  Octinion uses a soft gripper to grip the strawberry fruit body and imposes a rotational motion while pulling the strawberry to detach it from the peduncle~\cite{de2018development}. The end-effectors above make either a gripping contact with the fruit body or with the peduncle during the harvesting action. But the end-effector reported in \cite{xiong2018design} doesn't grip the fruit body or the peduncle during the harvesting action. Instead, a combination of three active and passive fingers guides the strawberry into the end-effector housing. Once the fruit reaches the cutting location, scissor-shaped blades cut the peduncle to detach the strawberry.

After the detaching operation, the end effector continues to catch/hold the fruit until it is dropped intentionally or safely placed in the designated location by the manipulator. Some end-effectors were reported to have certain finger arrangements to perform the catching action when the harvested fruits were dropped after the fruit detachment action. One such catching provision was provided in the end effector design reported by Arad B et.al.~\cite{arad2020development} for the sweet pepper harvesting robot. It was a soft plastic coated six metallic fingers arrangement just below the cutting blade assembly, that receives the fruit after the detachment. Another catching mechanism was proposed by Davidson J et.al.~\cite{davidson2017dual} for the apple harvesting robot. It used a two DOF secondary mechanism with a funnel-like catching end effector which will be moved to the dropping position to catch the apple while the primary picking manipulator detaches and drops the apple. This pick and catch approach was determined to be superior to the conventional pick and place approaches as it resulted in a fifty per cent reduction in the harvesting cycle time~\cite{davidson2017dual}.

Considering the different options for gripping and cutting, it is always beneficial to avoid applying any force on the fruit body by the end-effector contact surfaces. Since some fruits are very soft and delicate, there are higher chances of bruising during such operations. Aliasgarian et al.~\cite{aliasgarian2013mechanical} showed strawberry fruits are more damaged when exposed to compression forces on their body. Hence, from the end-effector design point of view, it is recommended to target the peduncle for gripping/cutting actions or to avoid a grip action as demonstrated in \cite{xiong2018design}.

\subsection{Perception} 

From traditional computer vision (CV) based to modern state-of-the-art deep neural networks, various methods exist for fruit detection and localisation of picking points. Traditional or classical CV approaches are typically based on geometric, thresholding, colour, and morphology. Thus, similar to other areas of CV, researchers have taken advantage of Deep Learning (DL) methods for performance improvement. 
Some of the early methods for detecting ripe strawberries relied on colour thresholding in HSI colour map~\cite{rajendra2009machine}. The authors also used thresholding of diameter for detecting the strawberry stem. The automatic thresholding-based algorithm was shown to be more robust by Zhuang et al. ~\cite{zhuang2019computer}. colour-based segmentation was used by Arefi et al.~\cite{arefi2011recognition} to segregate the background from the fruit blob.

Arefi et al.~\cite{arefi2011recognition} used colour-based segmentation to remove the background and keep the fruit blob. Instead of directly using colour, colour information can also be used with other features for a more robust approach. 3D-parametric model-fitting was used for the localisation of sweet peppers by Lehner et al.~\cite{lehnert2017autonomous}. Tao et al.~\cite{tao2017automatic} used geometric features with GA-SVM for apple classification. Arefi et al.~\cite{arefi2011recognition} combined the water-shed algorithm to extract the morphology of tomatoes from colour-thresholded binary images. Zhuang et al.~\cite{zhuang2019computer} were able to improve the results obtained by colour segmentation by using iterative-retinax algorithm along with Otsu's thresholding. 
Similarly, geometry-based algorithms are among the early contributions in this domain. Li et al.~\cite{li2020detection} applied morphological operations for litchi harvesting. The connected component algorithm was used by Duran et al.~\cite{durand2017real} to identify strawberry blobs. Moving on to geometry-based approaches, Hayashi et al.~\cite{hayashi2010evaluation} relied on extracted geometric features of strawberries to calculate stem angle w.r.t. to the longitudinal axis. \cite{tao2017automatic} used a Fast Point Feature (FPF) histogram, which is a geometric descriptor. The FPF descriptor consists of the parameterised query of the spatial differences between a point and its adjacent area which helps in describing the geometric properties within the K-neighbourhood of the point.

While threshold, colour, morphology, and geometry-based methods may provide good performance, they lack generalisation and are prone to noise. This is especially true for fruits like strawberries which are not regular in shape and lack symmetry. To improve generalisation, researchers need to engineer handcrafted features. However, with increasing size and variation in datasets handcrafting features become infeasible \cite{o2019deep}. The alternative is to use DL methods which are reviewed next. The most obvious DL-based approach is to use CNNs. Liu et al.~\cite{liu2018robust} combined CNN-based fruit detection with depth data to localise the fruit in 3D. CNN-based model for strawberry detection was also used by Lamb et al.~\cite{lamb2018strawberry} where the network was optimised through image tiling, input compression, network compression, and colour masking. Zhang et al.~\cite{zhang2018deep} relied on their CNN-based model for tomato classification. Instead of using colour images, Gao et al.~\cite{gao2020real} relied on a spectral features-based CNN model for detecting the ripeness and quality of strawberries. Thermal images have also been used as input to a CNN-based model for bruise detection on pears \cite{zeng2020detection}.
In recent years DL has been demonstrated to be superior for tasks such as segmentation \cite{he2017mask} and key-points detection \cite{cao2019openpose}. Thus, authors in selective harvesting have begun to adopt some of the DL techniques for fruit perception.

Lamb et al.~\cite{lamb2018strawberry} used CNN for strawberry detection by optimising the network through input compression, image tiling, colour masking, and network compression. Liu et al.~\cite{liu2018robust} used CNN in combination with depth data to calculate the relative 3-D location of fruit. Similarly, Zhang et al.~\cite{zhang2018deep} used CNN for tomato classification. Spectral features with CNN are used for strawberry quality or ripeness detection~\cite{gao2020real}. CNN model is also used for pear bruise detection based on thermal images \cite{zeng2020detection}. CNNs have revolutionised object detection and recognition, however for pixel-wise tasks such as semantic segmentation, Regional CNNs (RCNNs) is more appropriate. 
Sa et al.~\cite{sa2016deepfruits} relied on a faster RCNN model for bounding box detection of fruit while fusing faster RCNN, RGB, and Infrared (IR) images.

More recently, Mask-RCNN \cite{he2017mask} has been presented as a better alternative to the original RCNN. In selective harvesting also Mask-RCNN has been shown to provide a higher degree of accuracy while performing a pixel-wise segmentation~\cite{ge2019fruit}. Liu et al.~\cite{liu2018visual} relied on both YOLOv3 and mask-RCNN (M-RCNN) for bounding box detection of citrus fruit. M-RCNN with the ResNet-150 as backbone provided better performance than YOLO-v3. Similarly, Perez et al.~\cite{perez2020fast} relied on M-RCNN for strawberry segmentation for harvesting. Yu et al.~\cite{yu2019fruit} presented another instance of M-RCNN used for selective harvesting where features from M-RCNN were used to determine the strawberry shapes. Afterwards, a geometrical algorithm was used to localise the strawberry picking point. Researchers have extracted features from R-CNNs and combined them with their own algorithm to improve the localisation of picking points \cite{ge2019fruit,liu2019cucumber}. Ge et al.~\cite{ge2019fruit} first used M-RCNN to determine strawberry pixels then, the extracted strawberry pixels were combined with depth data, and thereafter density-based clustering and Hough transformation were used to develop a richer scene segmentation. Liu et al.~\cite{liu2019cucumber} combined M-RCNN with the logical green operator to come up with a more robust cucumber detection. Ganesh et al.~\cite{ganesh2019deep} used both HSV and RGB images to enhance the performance of M-RCNN for Orange detection. Yu et al.~\cite{yu2019fruit} applied M-RCNN to segment strawberry images and then used geometrical features to localise the picking point. 

On the other hand, Tafuro et al. \cite{tafuro2022} argue that localisation of picking points is not feasible by geometrical, statistical, or other such approaches even after fruit segmentation through M-RCNN. Instead, the authors rely on key-point detection normally used for tasks like human pose estimation \cite{cao2019openpose}, and face pose estimation \cite{zhang2014facial} for localisation of picking points.

We rely on the approach by Tafuro et al. for the initial detection and segmentation of strawberries. However, 2D detection is not sufficient for strawberry harvesting in 3D. As discussed earlier, the depth information is not sufficiently accurate owing to sensor inaccuracies and sunlight. Moreover, in the real-world more inaccuracies are introduced by camera calibration errors. Thus, it is not feasible to simply combine the 2D localisation with depth information. Similar, to Ge et al.~\cite{ge2019fruit}, we develop our own algorithm to refine and work around the inaccurate depth information obtained from depth sensors. However, our advantages over Ge et al.~\cite{ge2019fruit} and other methods based on M-RCNN discussed above are two-fold: 1) We rely on M-RCNN-based key-point detection for picking points which gives us much more robust picking point localisation as compared to handcrafted methods adopted to refine Mask-RCNN output.  2) We compensate for the lack of precise depth information by carefully fine-tuning the end-effector pose by translating the information to two additional cameras at the front.

\section{Concept, Design, and Features}
\label{sec:consept}

\subsection{System overview}

\begin{figure}[tb!]
\centering
  \begin{subfigure}[b]{0.36\textwidth}
    \includegraphics[width=\textwidth]{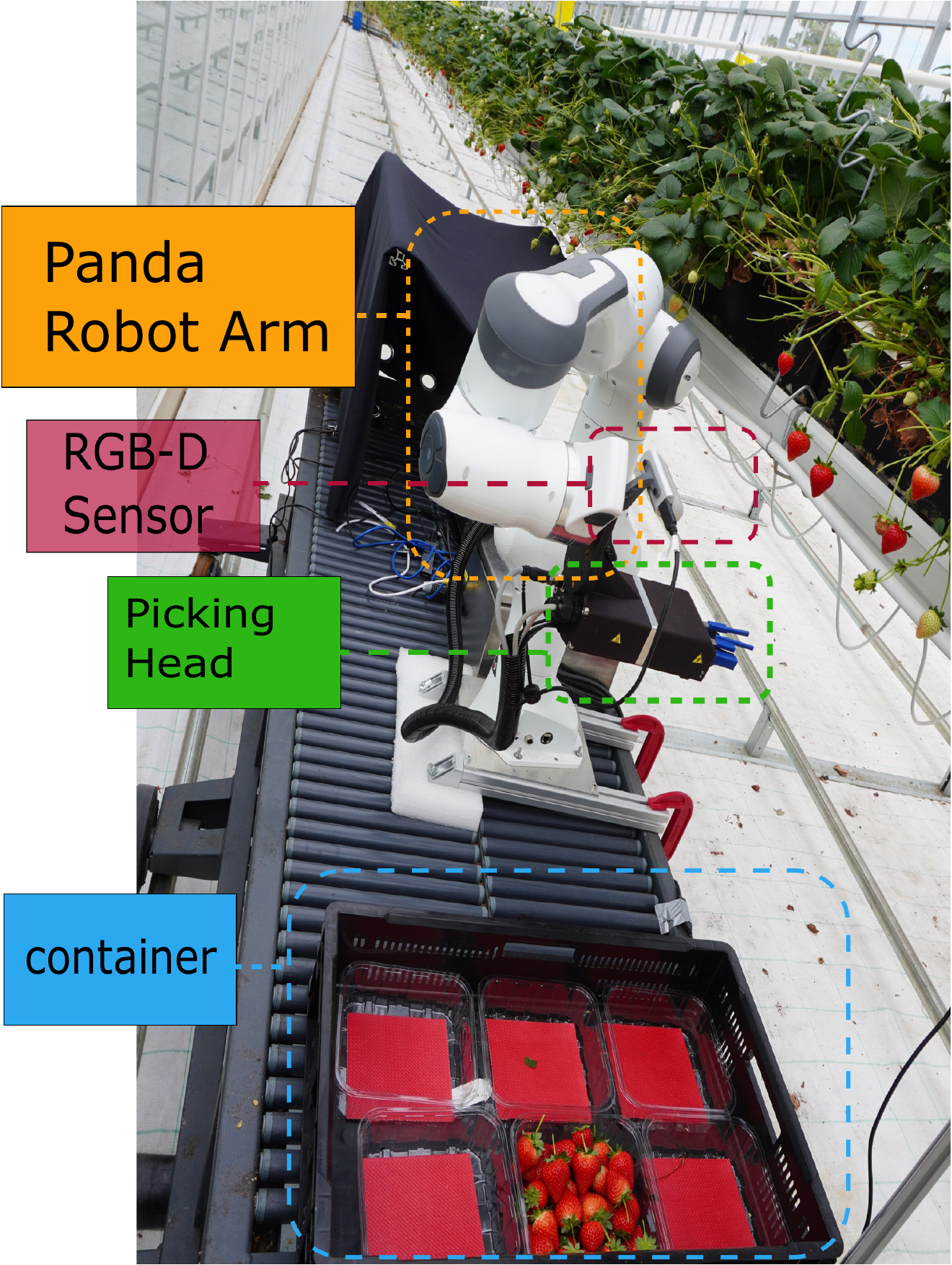}
    \caption{}
    \label{fig:field_test}
  \end{subfigure}
  \begin{subfigure}[b]{0.3\textwidth}
    \includegraphics[width=\textwidth]{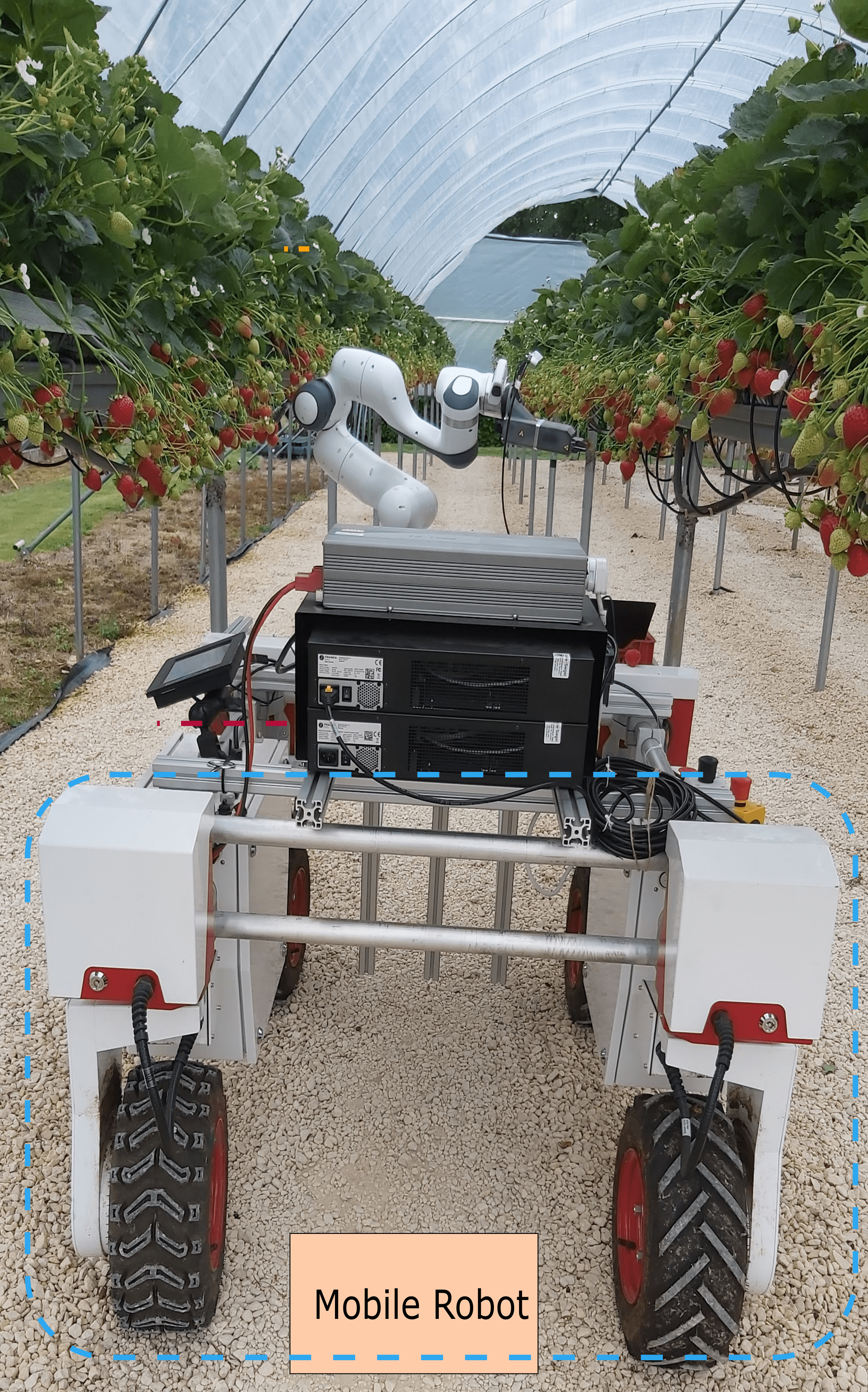}
    \caption{}
    \label{fig:thorvald}
  \end{subfigure}
  \caption{a) The system compromises a robotic arm (Franka Emika Panda), a designed picking head, an RGB-D sensor, and a fruit container. The system and its components including controllers are mounted on the commercial trolley capable of moving on a rail between the strawberry rows. b) All components of the system were reconfigured and mounted on a commercial mobile robot to be tested in a different strawberry growing field.}
  \label{fig:figure1}
\end{figure}

\begin{figure}[ht]
  \includegraphics[width=1\textwidth]{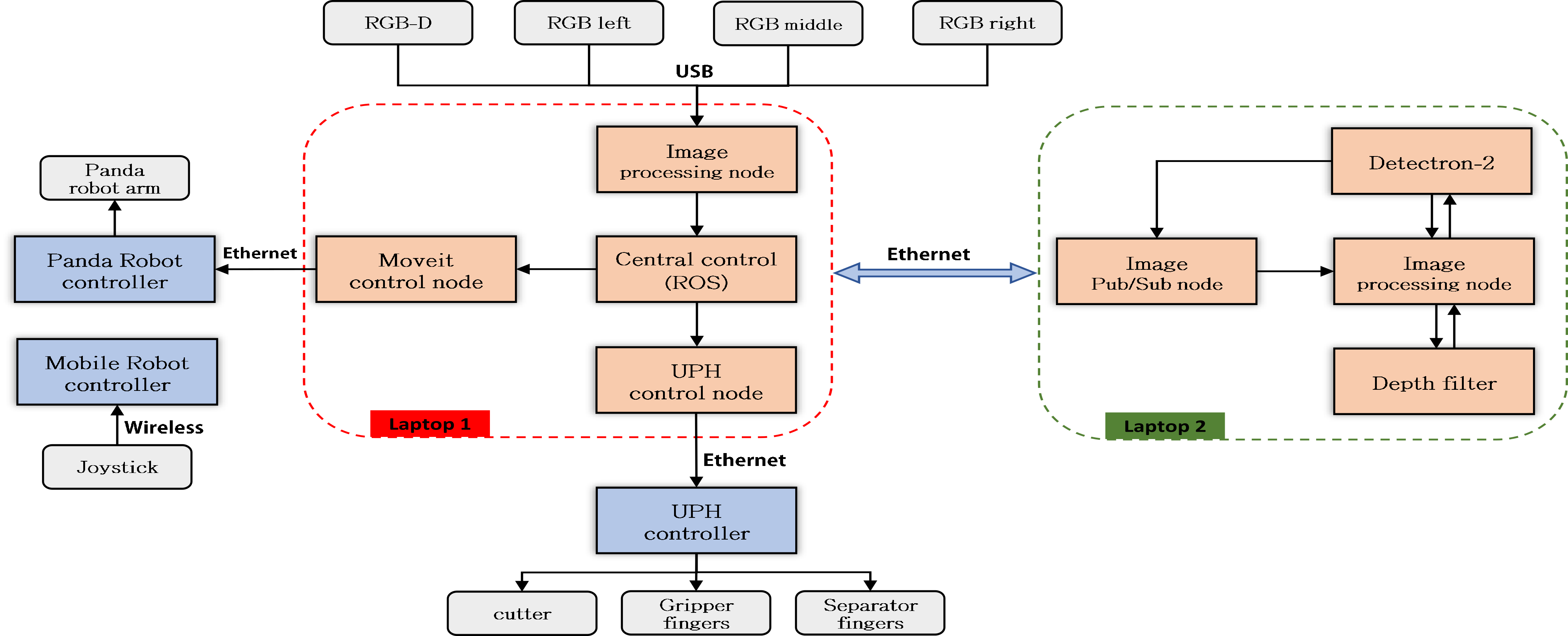}
  \centering
  \caption{System block diagram demonstrating all sub-systems and other components including; the robotic arm, mobile platform, perception, and control system.}
  \label{fig:block_diagram}
  \vspace{-4mm}
\end{figure}

To address the current challenges in the autonomous selective harvesting sector an autonomous system for fruit harvesting was designed and developed. The system includes a robotic arm, a novel robotic end-effector designed and manufactured for this research, a comprehensive perception system, a mobile platform; and an integrated control system for controlling the robotic arm and end-effector. Figures~\ref{fig:field_test} and \ref{fig:thorvald} show the autonomous fruit picking system and its components during field tests in a commercial strawberry growing glasshouse and research strawberry poly tunnels respectively. The robotic arm used in this work is an off-the-shelf arm Franka Emika Panda with a 3 kg payload and 7 degrees of freedom. A block diagram of the system is shown in Figure~\ref{fig:block_diagram}.
In this work, a novel universal picking head (UPH) for fruit harvesting was designed and introduced. The design of this picking head was to address the shortcoming of the available solutions, specifically for harvesting in dense clusters. We designed, manufactured, and successfully tested the picking head which has 2.5 degrees of freedom to allow harvesting fruits and manipulating possible occlusions independently. In the next sections, the design of the picking head is discussed in detail.

Our comprehensive perception system includes an RGB-D sensor and three RGB cameras, a novel dataset, and state-of-the-art algorithms to detect and localise the fruit and determine its suitability for picking. The RGB-D sensor is an Intel Realsense D435i model which is integrated into the picking head design and works based on eye-to-hand principles. This sensor provides an RGB image of the plant and also a three-dimensional point cloud to detect fruit, localise the picking point and predict the ripeness of the fruit. The RGB cameras located underneath the UPH allow close-range view where the RGB-D sensor loses its view. These sensors are coupled with a novel Mask-RCNN-based algorithm to form the perception system which is discussed in detail.

The robotic arm controller, UPH controller, RGB-D sensor, and RGB cameras are all connected to a laptop using either USB or CAN to USB adaptors. The laptop has an Intel Core i7® CPU, with 16 GB ram, and runs on Ubuntu 20.04.4 LTS (Focal Fossa). We used ROS Noetic (Robotics Operating System) as a middleware operating system to integrate all algorithms, sensors, robotic arm, and UPH and establish communication between them. The system includes a second laptop with a powerful GPU unit (NVIDIA® GeForce® RTX 2070 SUPER) to Handel the high-demand perception tasks. The second laptop is connected to the first laptop using an Ethernet connection and communicating through ROS. The system including the robot, controllers, and other components can be powered by a domestic power plug or provided by a DC power source e.g. a battery. In our field test, we tested the system using both methods. The DC power was sourced from the mobile robot batteries. The system also includes a container to hold the fruit punnets. The container could be replaced manually after it is filled. A rough estimate of the cost of the entire system (including the robotic arm, UPH, sensors, and computing system) is around £25k.

\subsection{A configurable and modular system}

Full integration of the system allows the robot to continuously detect and harvest ripe strawberries along the table rows. Similar to the other agricultural robotic harvesters, all sequences are performed in a static condition, i.e. when a strawberry is detected the mobile platform stops, the system harvests all reachable strawberries and the mobile platform moves on. The block diagram for the system includes all sub-systems and components shown in figure~\ref{fig:block_diagram} demonstrating the hardware and software architecture of the system.

One of the features of our system in comparison with previous works is the modular design of the system. As can be seen, all main components i.e. robotic arm, UPH, perception, and mobile platform are independent and could be integrated with other models with minimal development requirements. For instance, it is possible to use a variety of off-the-shelf robotic arms depending on the costs and the harvesting condition such as indoor/outdoor, task space, etc. More importantly, the system is configurable. In other words, the system can be configured for different growing conditions such as on-the-rail for glasshouses, on a mobile robot for full autonomous harvesting, or on an XY gantry mechanism for vertical farming.       

For this work, we integrated and tested our system on two mobile platforms in different conditions. First, the developed system including the robotic arm, UPH, and other components was mounted on Thorvald, a four-wheeled mobile robot made by Saga Robotics as shown in Figure~\ref{fig:thorvald}. In this case, the mobile robot was teleoperated remotely using a joystick. The harvesting test was carried out at strawberry-growing polytunnels at the Riseholme campus of the University of Lincoln. This strawberry-growing facility was established for research purposes. 

The second layout was a commercial strawberry harvesting trolley mounted on rails which was able to move between strawberry rows. In the current stage of this work, the trolley is operated manually, i.e. a human operator pushed the trolley on the rail. However, the automation of the mobile platform on the rail requires a simple solution that is widely available and out of the scope of this work. This field test was performed at the Dyson glasshouse strawberry growing facility which is a leading commercial strawberry grower in the UK as shown in Figure~\ref{fig:field_test}.

In addition to the hardware modularity, the software and control platform of the system is modular and configurable. The different units and control nodes are shown in Figure~\ref{fig:block_diagram}. These units include sensor data acquisition and processing, UPH control, robotic arm control, central control unit, a detection unit, and a ripeness assessor. For this project, we used two laptops to separate computationally heavy fruit detection which requires a powerful GPU, and robot control which requires high frequency. For both laptops, an image processing node is required to provide the correct format as different third-party packages are running on both laptops. These units could be developed independently and integrated depending on the harvesting condition, used hardware, and other requirements. As most available harvesting technologies were developed for a specific growing condition or strawberry variety, they are commercially unavailable. In addition, while they might have good performance for the condition they were designed for, they perform poorly for different conditions. This is important as the high number of combinations of different varieties and different growing conditions makes it impossible to propose a robotic solution for all of them. 

\subsection{System work-flow and algorithm}

\begin{figure}[tb!]
\begin{center}
  \includegraphics[width=1\linewidth]{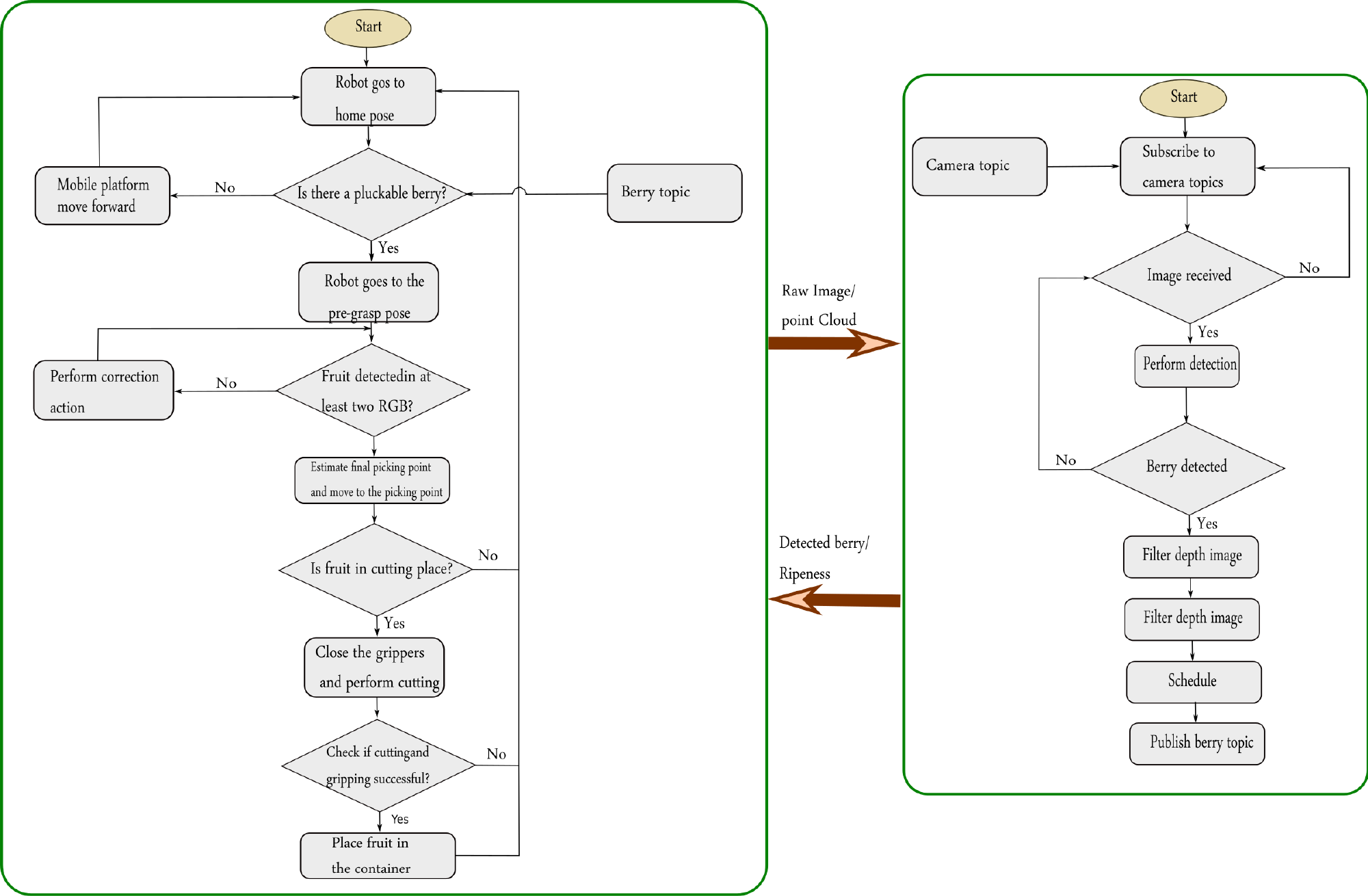}
\end{center}
  \caption{Flowchart of the perception and workflow of the system. The algorithm is implemented in two laptops, one running the ROS nodes/topics and high-level controller and the other a GPU-enabled system for running the strawberry detection and scheduling system.}
\label{fig:flowchart}
\end{figure}

The flowchart of the system algorithm and workflow is shown in Figure~\ref{fig:flowchart}. As can be seen, the entire algorithm was implemented in two laptops communicating through Ethernet protocol. The whole system consists of several control loops on different levels, i.e. high-level, mid-level, and low-level. Initially, the main loop triggers the robot control loop and perception loop. The robot goes to the home position (i.e., a suitable configuration in which the robot looks at the tabletop strawberries as shown in Fig.~\ref{fig:figure1}) and perception acquires the data, i.e. images, depths, point cloud, from multiple sensors. After pre-processing the received data, the perception algorithm detects all strawberries in the field of view and publishes their coordinates through the berry topic. In addition, the perception classifies all detected berries as "pluckable" or "unpluckable" which is included in the berry topic. 

Receiving an "at least one pluckable berry detected" message, triggers the robot control loop which converts the coordinates of berries from the camera frame to the robot base frame and commands the robot end-effector to move to the pre-grasp pose. As transformed coordinates of the picking points of the Strawberries in the robot base frame always contain some level of error, instead of the robot moving directly to the picking point pose, it goes to the pre-grasp pose. The pre-grasp pose is a point with the actual $Y$ and $Z$ coordinate and $X-d$ in the robot base frame where $d$ is a predefined distance of the end-effector from the berry. If more than one pluckable berries are detected the scheduling algorithm is triggered to determine the sequence of picking in order to improve the efficiency of the system.  

In the pre-grasp pose, the targeted berry is detected with at least two of the bottom RGB cameras. Using the detected coordinates of the targeted berry in the bottom RGB cameras, the errors of the picking point coordinates are estimated. Using the estimated errors, the picking point coordinates are corrected and the robot moves to the picking pose. in this stage, the strawberry stem should be located in between the gripping fingers where the cutter is able to cut the stem effectively. In addition, the remaining stem on the berry should not be too long which damages the other fruit in the punnet, or too short where the gripper fingers contact the fruit and bruise it. To confirm that the fruit is in the right place, a cutting confirmation algorithm based on the bottom RGB sensor was developed which is described in detail in Section~\ref{sec:cutting_command}.

After the confirmation that the fruit is located in the right place, the cutting command is sent and the stem is cut. In this stage, different conditions and scenarios can lead to unsuccessfully cutting and picking. To improve the efficiency of the system, and avoid redundant movement of the robot arm, a picking confirmation is performed before moving to place the fruit in the punnet. After the cutting action, the robot moves back with a pre-defined distance and performs a picking validation as outlined in Section~\ref{sec:cutting_command}.

If the picking is confirmed, the robot goes to punnet pose and places the picked berry in the punnet and the sequence is repeated. Placing poses of the fruit in the punnet are pre-defined points by which it is ensured that the fruits are placed evenly in order in the punnet to avoid bruising them.

\subsection{Control and Motion planning}

As the space between the strawberry plants and the robot arm is very limited and the environment contains a high level of uncertainty, there is a high possibility of collision of the robot arm or the end-effector with different objects. The high possibility of collision and limited task space demands a thorough and comprehensive motion planning approach. The manipulator's motion trajectory, velocity, and acceleration should be rectified from the beginning through all the way to putting the fruit in the punnet, including approaching the fruit, cutting action, holding, and placing.

In this work, to preserve a collision-free manipulation and prevent dropping/damaging fruit or equipment we defined a set of $n$ key points through the trajectory of the end effector motion denoted by $P_i$ where $i=\{0, 1, 2, ..., n\}$. We assumed that the trajectory of the mobile robot is aligned with the fruit tables, hence, the robot arm’s base frame is almost the same distance as the fruit tables. In this way, the robot arm’s home position always is almost the same distance as the fruit tables.  Different motion planning algorithms and trajectory/velocity/acceleration profiles were employed for each segment of movement to ensure collision-free and efficient manipulation.

At the beginning of the picking process, the end-effector frame is located at $P_0$, i.e. Home Position. The end-effector frame denoted by $\mathcal{F}_{ee} : \{O_{ee}; x_{ee}, y_{ee}, z_{ee}\}$ is attached to the middle of the gripper fingers. The end-effector frame coordinates are a defined point that coincides with defined trajectory key points during the robot's movement. When the $\mathcal{F}_{ee}$ locate at $P_0$, the system is initialised and the picking process is commenced.

We exploited the Open Motion Planning Library (OMPL)~\cite{ioan2012ieee} which is a sampling-based motion planning and Pilz Industrial Motion Planner to plan the manipulator's movement. Many planners in OMPL (including the default one) favour the speed of finding a solution path over path quality. A feasible path is smoothed and shortened in a post-processing stage to obtain a path that is closer to optimal. However, there is no guarantee that a global optimum is found or that the same solution is found each time since the algorithms in OMPL are probabilistic. Pilz Industrial Motion Planner provides a trajectory generator to plan standard robot motions like Pint to Point (PTP), Line (LIN), and Circle (CIRC) with the interface of a MoveIt planner. For this work, we used the LIN motion command. LIN motion planner connects the start point and end point with a straight line and generates a linear Cartesian trajectory between the goal and starts poses. The planner uses the Cartesian limits to generate a trapezoidal velocity profile in Cartesian space. This planner generates more accurate movement in Cartesian space with a focus on the end-effector trajectory.   

\begin{figure}[tb!]
\begin{center}
  \includegraphics[height=7cm, width=0.9\linewidth,clip]{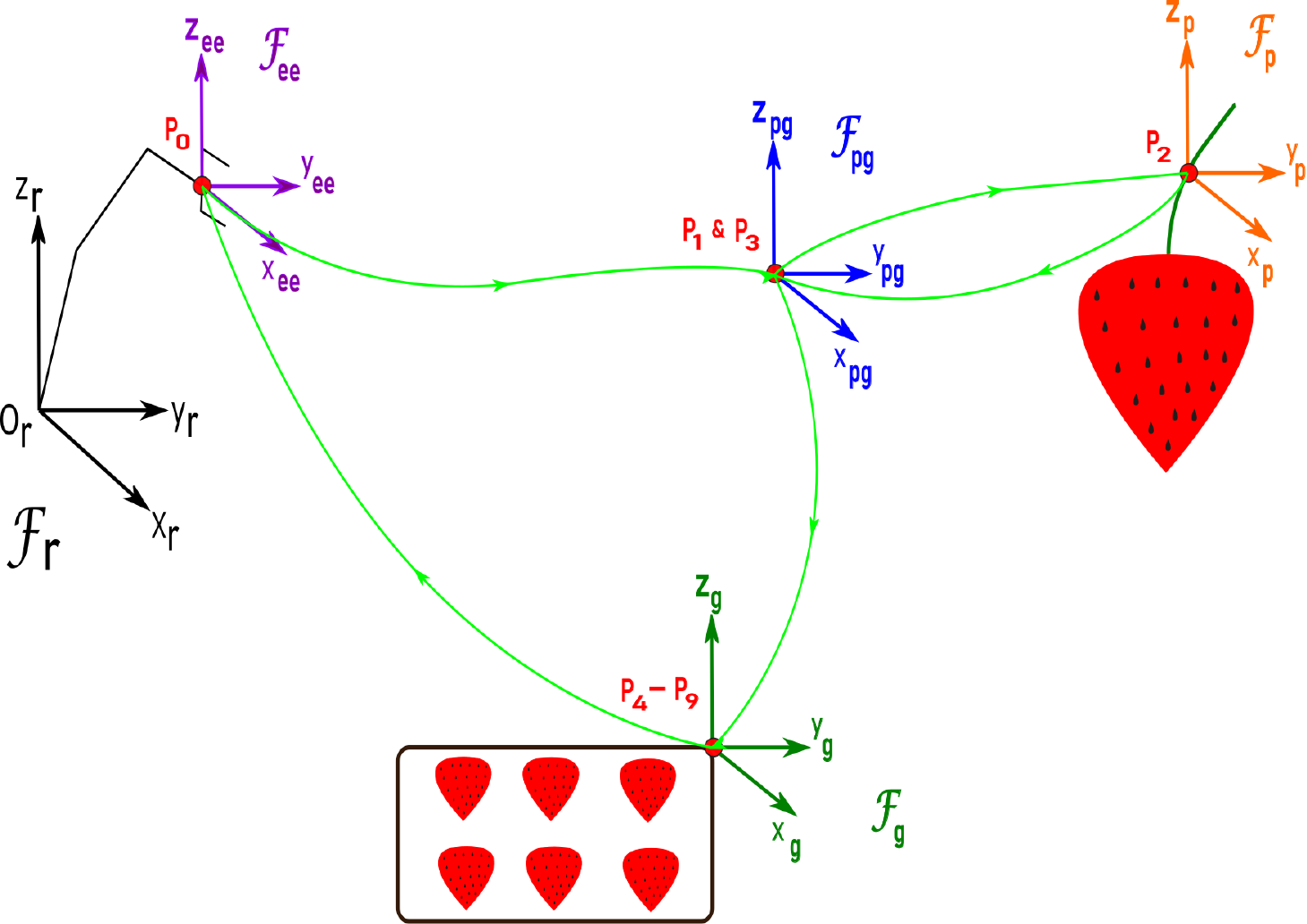}
\end{center}
  \caption{The initial ($t = 0$) and final state ($t = T$) of end-effector trajectory $\zeta(t)$ (shown with a green line): $\mathcal{F}_e$ is the robot's end-effector frame, the frame $\mathcal{F}_{pg}$ is attached to pre-grasp and post-grasp points, and a grasping configuration $\mathcal{F}_p$ is shown with a frame attached to the picking point. Also, a frame is attached to the container as the final goal of the end-effector is shown as $\mathcal{F}_g$. All frames are expressed using the inertial global frame $\mathcal{F}_r$.}
\label{fig:schematic}
\end{figure}

Figure \ref{fig:schematic} shows a schismatic diagram of the end effector trajectory $\zeta(t)$, attached frames, and key points. From $P_0$ to $P_1$, i.e. Grasp Pose, we used OMPL for motion planning as it doesn't require high accuracy movement nor a specific trajectory of movement. We set the speed at the highest level as long as preserves the safety and stability of the movement. For reach-to-grasp movement, $P_1$ to $P_2$, Pilz Industrial Motion Planner was employed to ensure the end effector goes through a defined trajectory to minimise disruption of the objects or possible collision. It is the same case for picking and validation action, from $P_2$ to $P_3$. For placing the strawberry in the punnet OMPL was used, however, the movement acceleration should be calculated carefully to avoid dropping the harvested strawberry. 

Assuming that the mass of the strawberry to be gripped in the end effector is about 50 grams. This is higher than the average mass value recorded during our field studies~\cite{vishnu2022peduncle}. For peak forces (F$_{C}$) about 22.53 N, coefficient of friction of 0.3, and safety factor of 2, under dynamic conditions, using Equation \ref{eq:acceleration} the end effector would be able to handle a 50 g strawberry for a manipulator acceleration of up to about 50 m/s$^2$.

\begin{equation}
F_c=\frac{m.(g+a).S}{\mu}
\label{eq:acceleration}
\end{equation}

where; 'F$_c$' is the maximum gripping force (N), 'm' is the mass to be handled (Kg), 'g' is the acceleration due to gravity (m/s$^2$), '$\mu$' is the coefficient of friction, and 'S' is a factor of safety. 

To place the harvested fruit in the punnet six key points were defined with respect to the frame attached to the punnet $\mathcal{F}_{g}$. These points were selected in a way that distributes the fruits in the punnet evenly and avoids bruising or damaging them. From placing point to the home position, again OMPL with the highest possible speed was used to increase the efficiency of the system

\subsection{Harvesting sequence planning}
\label{sec:scheduled}
selecting a berry as the target for picking among many possible pickable fruits is important to increase the success rate and reduce possible occlusions. Picking a free fruit that occludes other pickable fruits, not only normally has a higher success rate, but also removes the possible occlusion. There are notable studies on sequence planning for robotic harvesting that proposed near-optimal solutions~\cite{kurtser2020elsevier}. In this work we mainly implemented and tested two methods for simplicity and reducing the amount of computation to determine the target berry as described in the followings:
\bigskip
\\
\emph{1- Target Berry Selection Using Min/Max:}
\newline
The min-max algorithm attempts to find the maximum of the minimum distances among all the bounding boxes of the detected berries. Considering that there are $N$ berries detected in the top camera image view, the minimum of all the distances between each berry with respect to all other detected berries are calculated as:
\begin{gather}
 d_i^{min} = min( d_{i,1}, d_{i,2} \dots d_{i,j} )   ;   \forall i \neq j 
\end{gather}
Then the maximum of these minimum distances is taken as follows: 
\begin{gather}
     d_{target} = max( d_1^{min}, d_2^{min} \dots d_N^{min}, ) 
\end{gather}
where $d_{target}$ represents the most isolated berry in a cluster which is scheduled to be harvested first. This ensures that the difficult-to-reach berries are harvested later which aids in avoiding damage to other berries while trying to reach the difficult ones. 
\bigskip
\\
\emph{2- Coordinate based sorting:}
\newline
Although the previous method gives a systematic tool to select the target berry, it was noticed that it is more complicated to achieve a reasonable success rate. One reason is that by changing the viewpoint of the RGB-D camera, the targeted berry might change without a change in reality. The more efficient and practical approach could be sorting berries based on their coordinate in the image frame e.g. left-to-right or right-to-left depending on the direction of moving of the mobile platform. In this way, the mobile platform doesn't have to go forward and backwards to harvest all berries.

\section{Universal picking head}
\label{sec:endeffector}
\subsection{concepts and requirement of a universal picking head}

Picking fruits in different growing conditions is a challenging problem for
selective harvesting technology, as it is difficult to design and build an effective
robotic device (known as an end-effector or picking head) that is able to deal
with complex picking operations. A human’s hand enables dexterous manipulation
of fruit with 27 degrees of freedom, and over 80 per cent of the grasping information can be encoded into just 6 Eigen grasps. In contrast, conventional robotic end-effectors are customised for specific applications, such as pick-and-place
operations in industrial environments~\cite{Jarrasse2014nero}.

Currently, there are two types of picking heads available for robotic
harvesting of high-value crops: (i) a picking head having a parallel jaw gripper,
which may not be suitable for all types of crops, and (ii) a picking head that has a
customised design for picking particular fruit in a very specific picking scenario, which is only suitable for a specific type of crop or method of harvesting.
Consequently, the effectiveness of commonly available robotic picking heads is
limited, as different robotic picking heads may be needed for different crop types.
Some robotic picking heads are used to pick soft fruits such as strawberries.
Some of the robotic picking heads that are currently available for picking
strawberries are cup-shaped picking heads, which have opening parts that locate the peduncle of a strawberry and position the strawberry in front of cutting scissors in order to harvest the strawberry. The cutting action causes the strawberry to detach from the plant and fall into a punnet for collecting the strawberries. In this example, the picking head does not directly touch the flesh of the strawberry, which minimises bruising. However, as the strawberry falls from a height into the punnet, the harvesting can inadvertently cause damage/bruising to the
fruit.

Furthermore, fruit placement within the punnet is not controlled, which may
result in uneven distribution of the fruit in the punnet (which may also cause damage to fruit that are below other fruit).
Similarly, the design of the cup-shaped picking head, and the design of
other types of picking heads, may not be suitable for harvesting crops that grow in dense clusters. 
The present design has therefore identified the need for an improved
apparatus for the automatic detection, selection, and harvesting of crops that grow in dense clusters.

\subsection{Peduncle gripping and cutting force for end-effector design}

To develop an end-effector solution that attempts to detach strawberries by targeting the peduncle, it is essential to understand certain physical properties of the peduncle. This includes the estimate of the required cutting force and the gripping force that can be applied to the peduncle. Knowing the cutting force while using a particular cutting blade profile, gives insight into a better selection of actuation systems to provide the required cutting force. Moreover, the practice of using off-the-shelf blades takes away the need of investing effort to design optimum blade profiles. Rather such blades can be directly used or can be custom-made according to the standard profile. If the end-effector is designed in such a way as to use interchangeable blades, replacing the blades would be easier during worn-out situations. Hence it is wise to use cutting blades with a standard profile and in an interchangeable configuration in the end-effectors. For this work, a comprehensive study on the gripping and cutting forces of strawberry peduncle was carried out \cite{vishnu2022peduncle}.

This study intended to estimate the limit of the gripping force that can be applied to the strawberry peduncle without crushing it. To understand this force limit, experiments were conducted by applying compression force (analogous to the gripping force) to the peduncle specimens using a Universal Testing Machine (UTM). The peduncles of ripe strawberries of both varieties were selected for preparing the specimens. 15 specimens of each variety were prepared so that the peduncle was 10 mm in length and were trimmed at a distance of 10 mm from the top surface of the ripe strawberry fruit. This specimen measurement can simulate the situation where an end effector grips the peduncle within 10-20 mm from the top surface of the strawberry top surface during harvesting. The specimen diameter varied from 1.40 mm to 2.22 mm for Katrina with a mean and standard deviation of 1.75 mm and 0.24 mm. And for Zara, the diameter varied from 1.43 mm to 2.33 mm with a mean and standard deviation of 1.76 mm and 0.25mm.

In addition, we studied the force required to cut the peduncle of a ripe strawberry using a standard blade. And as an extension to this, the variation of this force at different cutting orientations was also studied. The profile of the selected cutting blade was studied using a scanning electron microscope. The blade had a double bevel cutting edge with a blade angle of 16.6$^0$ and a thickness of 0.22 mm. We studied the cutting force variation for 0$^0$,10$^0$, 20$^0$, 30$^0$ inclinations. 15 peduncle samples from both Zara and Katrina varieties were prepared for this study, i.e., 15 samples from each variety for each orientation of cut. These specimens were prepared by trimming the peduncle at a length of 30 mm from the top surface of the ripe strawberry.

During the experimental trials for studying the limit of the gripping force, all tested specimens showed a common force profile under compression load. While applying compression load to the specimen, there is a gradual increase in the resistive force to a certain point, and from then it shows a sudden drop. After then, the specimen gets squeezed completely on further application of compression load. It has been noticed that, after the drop in the resistive force, the specimen goes into permanent deformation, and finally leads toward complete squeezing. So this peak force (F$_{C}$) before the drop is considered the point of interest. The trend of this force (F$_{C}$) can be studied to limit the gripping force on the peduncle such that there is a lesser chance of squeezing the peduncle during the gripping action. The squeezing or crushing of the peduncle during the gripping action can result in the detached strawberry falling off from the grip during the harvesting process.

Considering the peak forces, we determined that the lowest of these peak forces (F$_{C}$) recorded is about 26.83 N and 22.53 N for Katrina and Zara respectively. This means that at these lowest values of compression force (analogous to gripping force), the respective test specimen went into permanent deformation before squeezing. Hence if we allocate a factor of safety of 2 to the lowest of these two values of forces (26.83 N and 22.53 N), the gripping force should be limited to around 10 N.

In addition, while analyzing the force values recorded during the cutting trials, again a common force profile has been noticed for all the tested specimens. In the profile, there is an increase in force value during the cutting action but with two sudden drops after two force peaks (F$_{P1}$ and F$_{P2}$). After the second peak force, there is a flat profile followed by a sharp rise in force after a point (E). This sharp increase happens when the blade touches the peduncle supports after cutting the peduncle off. So from the force profile for each specimen, the maximum of the two peak forces (F$_{P1}$ or F$_{P2}$) is taken as the peak cutting force (F$_P$) required for that specimen.

From the force values recorded at different cutting orientations, it has been noticed that the mean cutting force shows a relatively lowest value at 30$^0$ orientation compared to other studied orientations. At 30$^0$ cutting orientation, the maximum of the peak cutting force (F$_P$) recorded is about 7.20 N for Katrina, and 5.80 N for Zara. And hence, with a factor of safety 2, the cutting force requirement can be approximated to 15 N which could be considered sufficient to cut the strawberry peduncle at 30$^0$ orientation. Also, this force would be sufficient to handle other cutting orientations studied. We exploited these results to optimise the design of the end-effector and increase the harvesting success rate.

\subsection{End-effector design}
\begin{figure}[tb!]
  \begin{subfigure}[b]{0.45\textwidth}
    \includegraphics[width=\textwidth]{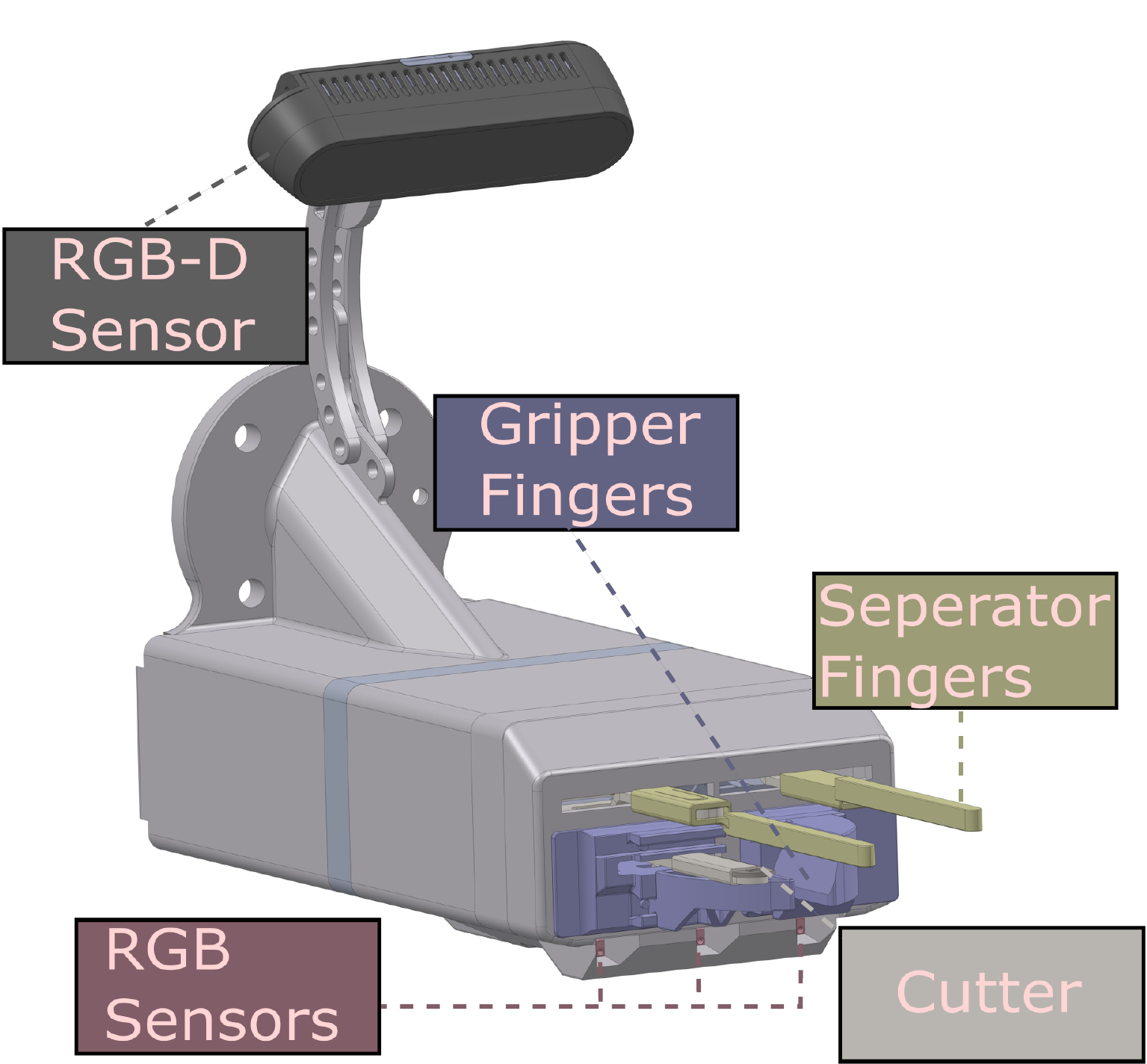}
    \caption{}
    \label{fig:design2}
  \end{subfigure}
  \hfill
  \begin{subfigure}[b]{0.45\textwidth}
    \includegraphics[width=\textwidth]{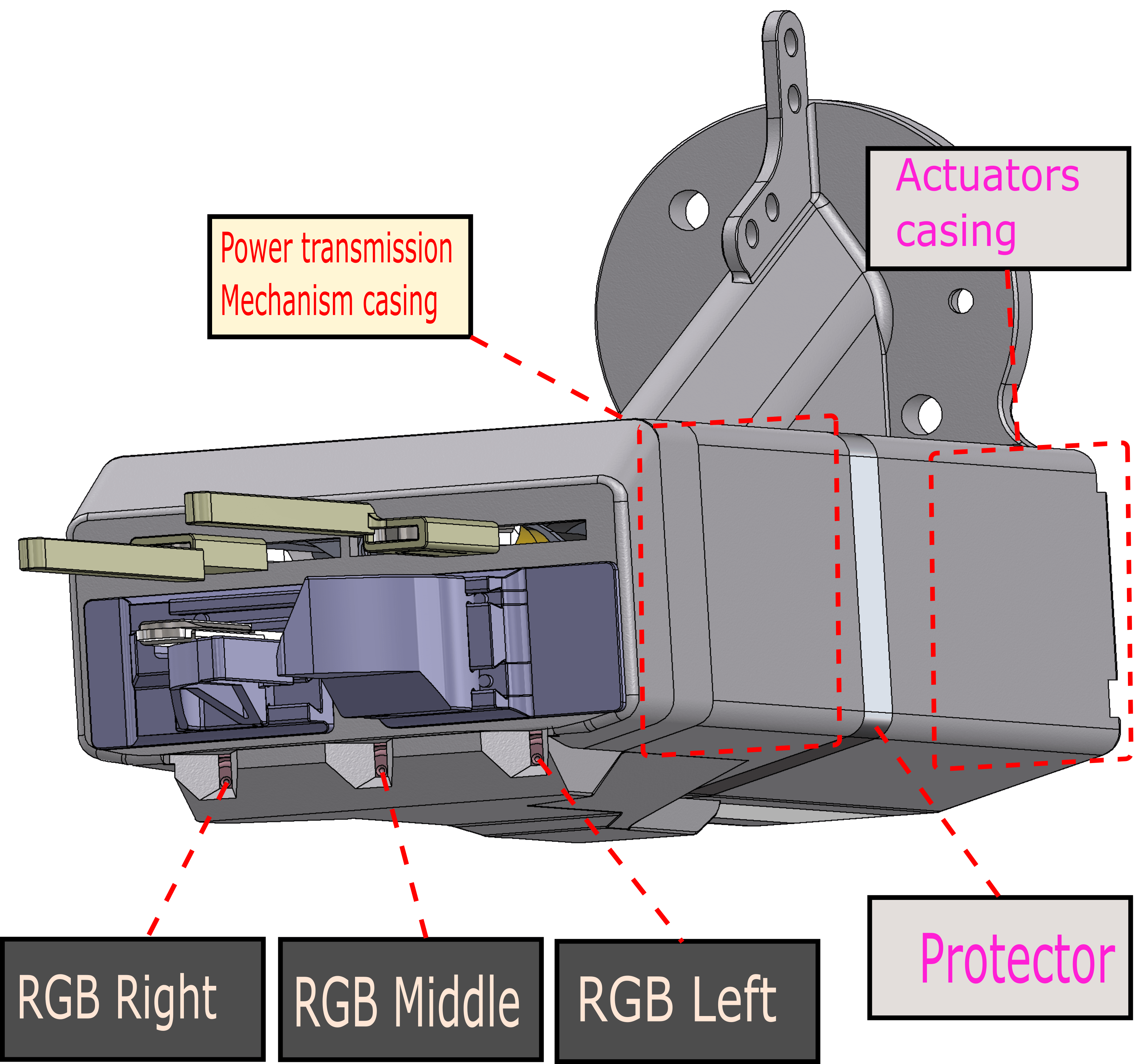}
    \caption{}
    \label{fig:design3}
  \end{subfigure}
  \caption{a) End-effector overall view and its components. The end-effector compromises an RGB-D sensor mounted on top of it for a better wide view, three RGB sensors for close-range view, the first pair of fingers (Separators) for occlusion removal, second pair of fingers for gripping and holding strawberry stem (grippers), and a cutting mechanism for cutting the stem (Cutter). b) End-effector internal design and components. It includes two independent actuators controlling grippers, separators, and cutter providing 2.5 degrees of freedom. The protector separates actuators from the power transmission mechanisms for better heat reduction and dust and water isolation.}
\end{figure}

The proposed robotic end-effector for fruit harvesting, comprising: a vision
system for identifying the location of the ripe fruits on the plant; the first pair of fingers for moving any objects that at least partly occlude the identified ripe fruit on the plant (Separators); the second pair of fingers for gripping a stem of the identified ripe fruit (Grippers); and a cutting mechanism for cutting the stem of the identified ripe fruit when the stem is gripped between the second pair of fingers (Cutter), wherein a portion of the stem that remains attached to the fruit remains gripped by the grippers after the stem has been cut. The design and components of the universal picking head are shown in Figures~\ref{fig:design2} and \ref{fig:design3}.

In general, the end-effector described herein benefits from 2.5 degrees of freedom, which is higher than those of available picking heads. This added degree of freedom allows the end-effector to deal with complex picking scenarios where the available picking heads fail. The end-effector benefits from an effective combination of actuation systems and sensors to resolve the limitations of currently available picking heads. The end-effector includes three separate movements (moving objects, gripping a stem, and cutting a stem) that are actuated using two actuators. This is useful as the ability of the end-effector is increased without also significantly increasing the complexity or component count of the device. The actuators and protector design for the picking head are shown in Figure~\ref{fig:design3}

The presented techniques advantageously enable ripe fruit to be harvested
without bruising or damaging the fruit. These techniques are particularly
advantageous for harvesting fruit that grows in dense clusters, such as
strawberries. Strawberry is an example -- and not limited -- fruit
that may be harvested using the robotic end-effector of the present techniques.
More generally, the presented techniques may be used to harvest different types of
fruit and vegetable crops, including those which grow individually and those which
grow in clusters.

The robotic end-effector comprises a first actuation mechanism
for controlling the actuation of the first pair of fingers. Thus, a dedicated actuation mechanism is used to control the movement and operation of the first pair of fingers which allows one degree of freedom for manipulating the cluster and removing possible occlusion independent from griping and holding the fruit.

Responsive for receiving the location of an object that at least partly occludes
a fruit, the first actuation mechanism controls the first pair of fingers to push
away the object, by increasing the separation distance between the first pair of
fingers. Thus, the first pair of fingers are closed together when the end-effector is being used to image a plant and identify ripe fruits, and/or when it is moving towards an identified ripe fruit. The first pair of
fingers are moved further apart when an object that at least partly occludes
fruit needs to be moved away so that the fruit can be better seen (to determine
if it is suitable for harvesting) and/or so that the second pair of fingers can grip the stem of the fruit.

The cutting mechanism is located in proximity to the second
pair of fingers, such that when the stem of the identified ripe fruit is cut by the cutting mechanism, the second pair of fingers continues to grip a portion of the stem that remains attached to the fruit. In other words, when a cutting operation performed by the cutting mechanism is complete, the fruit is not immediately dropped into a container for collecting the harvested fruit. Instead, the fruit continues to be gripped, via a portion of the stem that is still attached to the fruit, by the second pair of fingers. This is advantageous because the robotic end-effector may be controlled to gently place the harvested fruit in a container.

The second pair of fingers release their grip on the portion of the stem that
remains attached to the fruit when the end-effector is close to the container. The cutting mechanism may be partially or fully covered or encased for safety reasons, i.e. to avoid any risk of a human operator being able to come into contact with the cutting mechanism.

The second pair of fingers is moved by a second actuation mechanism which also moves the cutting mechanism. As the cutting mechanism is only operated when it is confirmed that the target fruit is located in between the second pair of fingers, advantageously a single actuation mechanism is used to control both the second pair of fingers and the cutting mechanism, thereby reducing complexity and the number of components needed to control the end-effector.

The vision system comprises a depth sensor for generating a three-dimensional
map of the plant. The vision system uses the three-dimensional map to identify the location of ripe fruits on a plant and any objects that at least partly occlude the identified ripe fruits. The depth sensor is an RGB-D (red-green-blue-depth) camera.

Furthermore, The vision system includes three RGB image sensors for capturing
images of the fruit/cluster of fruits at the bottom of the end effector in the vicinity of the second pair of fingers to enable a better view in close range. 
Having the RGB sensors in the vicinity of the second pair of fingers
is advantageous because the sensors capture images of the fruit or cluster
of fruits at the fruit level, whereas other sensors of the vision system may view the
fruit from a different perspective/angle. This also reduces the risk of every sensor
of the vision system being occluded during the picking process, i.e. it provides
some redundancy in the vision system.

The effective configuration of RGB and RGB-D sensors helps to efficiently
detect and localise the ripe fruits. In addition, the combined sensory
information can be used to estimate the size, weight, and sort quality of the
fruits to be picked. Also, they are further used to control fruit picking and occlusion removal actions.

\input{perception.tex}

\input{results.tex}

\section{Discussion}
\label{sec:descussion}
The results of the field experiments demonstrate the robustness and effectiveness of the new robotic harvesting system. The experiments in two different growing conditions and three different strawberry varieties show the high-level adaptability of the system. It proves that the modular characteristics of the system alongside its high ability to reconfigure the system based on the required condition, show distinguished performance. this ability is highly important to generalise its developed technologies to other fruits or growing conditions with minimal changes. The current alternatives have been designed and developed based on specific needs and conditions, which makes it challenging to adapt them to other environments. 

The vision system and ripeness detection proved to be effective with just 4.9\% of the ripe fruit not detected. From the graph shown in Figure~\ref{fig:error} it can be seen that the localisation of the vision system contains a level of significant errors, however, the errors fluctuate around a specific value. Some of these errors could be caused by the system's internal calibration errors. For instance, camera calibration errors, or errors of transformation from camera coordinate to robot coordinate can contribute to internal calibration errors. Another source of error could be the different lighting conditions of the experiment environment. Regardless of the source or nature of the errors, the proposed Gaussian Process Regression for picking point error estimation method proved to be effective of mitigate the position errors. From Table~\ref{table:performance} it can be seen that just 8.4\% of the attempts failed due to position error. It is notable that we used a limited dataset to train the model. A more comprehensive dataset with more data points can enhance the performance of the model in different conditions.

The cutting confirmation and picking validation methods were also tested during field experiments. This sub-system improves the efficiency of the whole system by reducing the redundant movement of the robot. However, the results show that around 12.9\% of all failed harvesting attempts were because of the cutting confirmation and picking validation failure. One reason for these failures that were observed during the field experiment lighting condition of the environment. As the strawberry, in general, has a shiny texture that reflects direct sunlight, in some scenarios the direction of the sunlight reflection prevents the RGB sensors to pick up and transmit the correct colour. In this situation, the pixels show bright colours instead of red colour which leads to classifying the picking as unsuccessful or the cutting command as not sent. 

Different factors affect the harvesting time of the robotic system. Figure~\ref{fig:time} shows that harvesting time varies from 21 seconds to 35 seconds with an average of 28.2 seconds. Field experiment observations indicate that one reason for time variation is robot correction movements to adjust to the targeted fruit location and/or bring the targeted fruit into the field view of the bottom RGB sensor. In some scenarios, the robot carries out multiple adjustment movements such as moving back, left, right, down, or up to bring the targeted fruit into the bottom sensors' field of view. These adjustment movements are important to capture the targeted fruit which is later used for validation, although it might increase the harvesting time slightly.      

The field experiments demonstrated the effectiveness of the novel design of the picking head. Only 4.4\% of all attempts failed because of the cutting or gripping failure. It means that the picking head design was effectively capable of a stable grip and successful cut of the strawberry stem. The picking head is successfully capable of grasping and manipulating the harvested strawberry without contact with the fruit. In contrast, most available technologies handle the fruit by grasping it using grippers or suction cups or directly dropping it into the container after cutting the stem. Our design grasps and handles the fruit from its stem reducing the possibility of bruising or damaging the fruit significantly. In addition, an ineffective cutting mechanism leads to partial cutting which damages the plant and increases plant disease possibility. We tested the design on different strawberry varieties which have different stem diameters and strengths. The experiments proved that the cutting mechanism is able to cut the different varieties of stems effectively. (A video of the system can be seen in \href{https://www.youtube.com/watch?v=JF4WR6Li-v4&ab_channel=UniversityofLincoln}{this link}.)

\section{Conclusion}


We designed, prototyped and field tested a novel picking head that can navigate through possible clusters and pick a targeted fruit. The picking head consists of two independent mechanisms for grasping and removing occlusion providing 2.5 DOF including the cutting mechanism. This novel design allows the system to manipulate occlusions independent of picking actions. In addition, the picking head design provides a contact-free grasping and picking of the fruit. This is highly important to reduce fruit damage or bruising and to reduce the corresponding waste as strawberries are a very delicate fruit. We developed and proposed a state-of-the-art perception system using RGB-D and three RGB sensors. To train our vision system, we produced two new datasets from a real strawberry growing farm with different features such as key points, picking points, ripeness, etc. We designed and developed the autonomous harvesting system modular and configurable to increase its adaptability for different strawberry varieties and growing conditions. Finally, to test the system we performed field experiments on two different commercial farms with three varieties. The field experiment results show the efficiency and reliability of the system with an 87\% success rate. Furthermore, the perception system demonstrated 95\% success in detecting ripe fruits.


\bibliographystyle{apalike}
\bibliography{egbib}

\vfill

\end{document}

%% file: perception.tex
\section{Perception for selective harvesting robotic system}
\label{sec:perception}

Strawberries are grown in dense clusters and they come in different configurations and wide varieties with varying shapes. Moreover, the asymmetric and irregular nature of the stems coming out of the fruit makes it difficult to localize the picking point. Commercially available depth sensors are designed for large objects under controlled lighting conditions. Insufficient quality of depth-sensing technologies makes strawberry picking point localization on stem intractable. Our depth sensing, namely RealSenhas 435i, which is widely used in robotics, has deteriorated performance under bright sunlight in farm conditions. In addition, it is designed to work best for distances larger than 50 [cm] where the preciseness drops to 0 for distances below 15 [cm]. As picking strawberries by grasping  them contributes to their bruising, we considered picking by griping and cutting the fruit stem requiring picking point (PP) localization. However, localizing the picking point in the depth image is challenging because of the low resolution of RealSense 435i, in particular  where the distance is below 15 [cm].  

Traditional methods rely on the color-based \cite{arefi2011recognition}, geometry-based, or shape-based \cite{li2020detection} methods.  Strawberries are neither very symmetrical nor their orientation is fixed making an accurate assumption, or shape-based prediction about their stem location difficult. Instead, we take advantage of the Deep Learning (DL)-based strawberry segmentation and key-point detection method proposed by Tafuro et al \cite{tafuro2022}. We use the key points to understand the orientation of the strawberry and pose the end-effector. The key point also localizes the picking point to a reasonable accuracy. However, due to the very thin cross-section of a strawberry stem, this cannot always exactly localize the stem giving rise to inaccuracies in the depth perception of the picking point. 

In addition, we address the lack of precise depth sensing and fine-tuning the localization in close proximity by a novel camera configuration (short and medium-distance focused cameras) and a combination of localization for short and medium distances. The system uses an Intel Realsense D435i depth-sensing camera as the main sensing system. The device is placed at the back and top of the end effector which allows sufficient distance from the cutting action for the depth sensor to be reasonably accurate. In addition to the Realsense camera, there are three RGB cameras placed at the front bottom of the end-effector. At this close range, it is also not feasible to calibrate the RGB cameras as a stereo system. Instead, we carefully find the project of 3D coordinates of the strawberry segmentation into 2D pixel coordinates of the front RGB cameras. Then, based on the coordinates of the strawberry bounding box visible in each image we trained an AI-Based model to make meticulously adjust the position of the end-effector to the final cutting position.
In addition to localizing the fruit and picking point, the perception should be able to determine the ripeness, size, and shape of the fruit and determine whether it is suitable for picking or not. we present two novel datasets of
strawberries annotated with picking points, key-points (such as the shoulder points, the contact point between the calyx and flesh, and the point on the flesh farthest from the calyx), and the weight and size of the berries. We performed experiments to predict if the fruit is suitable for picking or not.

In contrast to the existing works in which classic CV methods are used to determine picking points and suitability for picking, our approach includes SOTA MRCNN models. We collected two datasets to train our models: Dataset-1 is collected at a new \emph{15-acre} table-top strawberry glasshouse in Carrington, Lincolnshire, which is the latest addition to SOTA  Dyson Farming’s circular farming system \cite{dyson2021}; Dataset-2 has been derived from the Strawberry Digital Images (SDI) \cite{PEREZBORRERO2020105736}.  Dataset-1 is a novel dataset that presents strawberry dimensions, weights, suitability for picking, instance segmentation, and key-points for grasping and picking action. The main purpose of this dataset is to facilitate autonomous robotic strawberry picking.

For each strawberry, the dataset presents five different key-points: the picking point (PP), the top, and bottom points of the fruit, the left grasping point (LGP), and the right grasping point (LGP). While the PP indicates the position on the stem where the cutting action has to be performed, the left and right grasping points can provide a reference to the end effector for the grasping action. In addition, the dataset also contains annotation for instance segmentation for each of the strawberries. To determine the suitability of strawberries for harvesting each strawberry is labeled as `pluckable'--ready to be picked-- or `"unpluckable"'--not to be picked--. "unpluckable" strawberries include unripe, semi-, and over-ripe or rotten berries. The `pluckable' category includes strawberries that are nearly ripe and perfectly ripe. The dataset contains a set of 532 strawberry sets. Each set has three colors, depth, and point cloud data of the same strawberry cluster from different distances. The farthest image captures the entire cluster whereas the nearest image focuses on one target strawberry in the cluster. In total, this dataset includes 1588 strawberry images. All the images have been captured with Intel Realsense RGB-D sensor \emph{D435i}. 

Dataset-2 is an enhancement of the SDI dataset~\cite{PEREZBORRERO2020105736}. SDI dataset contains a total of 3100 images.  These are dense strawberry clusters that contain an average of 5.8 strawberries per image. We carefully annotated 10999 berries each with 5 different key-points. Moreover, we labeled "pluckability" (i.e. suitability to be picked) of all the strawberries. The strawberries not annotated for key-points are either severely occluded or are in an early flowering stage where a meaningful annotation is not possible. more details on the datasets and proposed methods were published previously and can be reached in~\cite{tafuro2022}.

\subsection{Segmentation, Key-points and Pluckable Detection}

Our proposed approach includes Detectron-2 \cite{wu2019detectron2} for segmentation and key-points estimation. The Detectron-2 model is based on MRCNN~\cite{he2017mask} and has become the standard for instance segmentation. It also has an added capability of key-points detection for human pose estimation. We adapted this key-points detection method integrated within Detectron-2 to estimate the strawberry key-points. The datasets' key-points, segmentation masks, and strawberry categories (`pluckable' and `"unpluckable"') are converted to MSCOCO JSON format~\cite{lin2015microsoft}. This MSCOCO JSON is the default format for feeding data into Detectron-2. It is also essential to recalculate the bounding box. Without the key-points, the bounding box aligns to the extremities of the segmentation mask. However, the PP key-point lies outside the segmentation mask of the strawberries and thus outside the bounding box. Because of the nature of MRCNN, a key-point outside the bounding box is not detectable. Thus, the bounding boxes are expanded to accommodate all the key-points. We performed experiments with three backbone networks for  Detectron-2, R50-FPN, X101, and X101-FPN. ResNeXt \cite{xie2017aggregated} (X101-FPN and X101) is  a more recent network that was introduced as an improvement to ResNet-50 (R50-FPN). Section \ref{sec:seg_kp_results} discusses the results in detail.

\subsection{Perception Setup} 

We use two laptops to control the entire system and run the perception system. The first laptop runs the ROS nodes as shown in Figure~\ref{fig:block_diagram} (left). The second laptop with Nvidia GPU runs the Detectron node as shown in Figure~\ref{fig:block_diagram} (right). The vision system consists of three cameras: An Intel Realsense d435i color and depth-sensing camera and three colors (RGB) cameras.  In robotic perception, depth-sensing cameras are essential for the 3D localization of the target and for generating a point cloud. However, most commercially available depth cameras including Realsense d435i are not suitable for close proximity sensing ($\leq15~cm$). Thus, for the depth sensing to be feasible the Realsense camera is mounted on the back and top of the UPH \ref{fig:design2}. However, due to the positioning of the Realsense, the berries are occluded by the picking head in close proximity to maneuvering.  So, to compensate for the lack of depth sensing and occlusion in close proximity maneuvering three RGB cameras are additionally mounted for fine-tuning the trajectory, cutting confirmation, and picking success validation. Considering that the purpose of these RGB cameras is error reduction instead of primary strawberry detection, these cameras were mounted firmly on the front bottom of the picking head with a 25 mm horizontal distance between them.

\subsection{Perception pipeline and approach}
\label{sec:percep_pipeline}

The berry plants are on both sides of a lane of strawberry farms with a tabletop system. Hence,  our robot, similar to human pickers, picks the ripe fruits on one table on one side before picking the fruits on the other side. This simplifies motion planning as collision is checked for one side of the row of tabletop strawberries. This was achieved using joint constraint in the robot arm planning.  Let $x_c^{top}$, $x_c^{left}$, $x_c^{middle}$, and $x_c^{right}$  be the coordinates of strawberry in the top (Intel RealSense), left, middle, and right color cameras respectively. Notice, $c \in {x,y,z}$ for $x_c^{top}$ i.e. 3D, $c \in {x,y}$ for $x_c^{right}$ and $x_c^{left}$ i.e 2D.  $\mathcal{P}^{top}$ is the camera projection matrix of the top camera for translating an image or pixel coordinates to camera link coordinates. This consists of camera intrinsic ($K^{top}$) and extrinsic ($R^{top}$, $t^{top}$) parameters, where $R$ and $t$ stands for rotation and translation respectively. $T^{top}$ is the camera calibration parameter/matrix which translates the camera link coordinates to the robot base frame. $K$ is the intrinsic parameters of the left, middle and right color camera. $T^{left}$, $T^{middle}$, and $T^{right}$ are the camera calibration parameters/matrices for projection from the robot base frame to the camera coordinate. The sequence of steps taken below for the picking action is shown in Figure \ref{fig:block_diagram} and described below:

\begin{itemize}

\item  At the home position, all strawberries are detected in the top (RealSense) camera image frame and scheduled. The depth estimation for the picking point is not reliable for something as thin as the strawberry stem. Thus the 2D segmented strawberry pixels are used as binary masks on the depth image to filter the depth pixels belonging to the strawberry. Further, depth pixels indicating depths less than 20cm and more than 50 cm are filtered out. The minimum distance of the gripping point from Real-Sense is roughly 20cm. Clearly, any strawberry pixel should be more than 20cm. Further, the initial position of the robot cannot be more than 50 cm from the berry due to the farm structure. Then we take the average value of the remaining strawberry depth pixels which gives us a more reliable 3D coordinate $x_c^{top}$ 

\item The scheduled berry coordinate is transformed from the top image or pixel coordinate to the robot base frame using the camera calibration matrix  

\begin{gather}
    \label{eq:pix2robot}
    X_c^{base} = \mathcal{T}_{base}^{top} \mathcal{P}^{top} x_c^{top} \\
    \label{eq:pix2camlink}
    where~{P}^{top} = [R^{top}, t^{top}]K^{top}
\end{gather}
Basically, the pixel coordinate is first transformed to the camera link coordinate using camera intrinsic and extrinsic parameters (Eq. \ref{eq:pix2camlink}) and then the camera link is transformed to the robot base frame (Eq. \ref{eq:pix2robot}).

\item Using careful calibration, the targeted berry coordinate $X_c^{base}$ in the top camera is back-projected to the bottom left, middle and right camera using the camera intrinsic ($K^{left}$, $K^{middle}$, $K^{right}$) and camera calibration parameters ($T^{left}, T^{middle}, T^{right}$).
\begin{gather}
    \label{eq:bp_left}
    x_c^{right'} = K^{right}\mathcal{T}_{left}^{base} X_c^{base}\\
    \label{eq:bp_middle}
    x_c^{middle'} = K^{middle}\mathcal{T}_{middle}^{base} X_c^{base}\\
    \label{eq:bp_right}
    x_c^{left'} = K^{left}\mathcal{T}_{right}^{base} X_c^{base}
\end{gather}

\item This back projection (Eqs. \ref{eq:bp_left}, \ref{eq:bp_middle} and \ref{eq:bp_right}) helps in establishing a target berry association in the top and bottom cameras at the robot home position based on image plane position error.
\begin{gather}
    \gamma^{left} = x_c^{left} -  x_c^{left'}\\
    \gamma^{middle} = x_c^{middle} -  x_c^{middle'}\\
    \gamma^{right} = x_c^{right} -  x_c^{right'}
\end{gather}

$\gamma$ is the error of the projected bounding box and the bounding boxes in the bottom cameras. We consider the berry with the lowest $\gamma$ the targeted berry in each camera view. The error arises mainly from inaccuracies in the depth perception of the RealSense camera as well as errors in camera calibration.

\item From this point on, for any arm movement, this berry association is identified in bottom cameras for later fine-tuning, grasp alignment, and cutting confirmation.

\item The UPH is then moved to a pre-grasp pose which is at a fixed distance $D$ from the berry. This means the arm is moved by  ($X_z^{base}$ - D) along the depth axis while aligned to the strawberry picking point in X and Y coordinates  ($X_{x^{'}y^{'}}^{base}$).

\item Once the arm is in the pre-grasp pose, the UPH gripping point is adjusted based on the estimated errors calculated. The goal here is to move the gripper to the picking point ($X_z^{base}$) based on initial depth sensing. However, the X and Y alignment is continuously fine-tuned, where the goal is to locate the grasp pose of the strawberry stem in between the gripper fingers.

Due to errors in depth sensing and camera calibration, the gripper cutting point does not align perfectly with the strawberry picking point. The novelty of the proposed method is that we back-project the 3D coordinate to 2D coordinates in the front color cameras. By assuming the fruit with the lowest $\gamma^{left}$ and $\gamma^{right}$ as the targeted fruit, we are able to associate the same berry with the bottom color cameras and fine-tune the X, and Y alignment based on more reliable strawberry coordinates in the bottom cameras. This enables the system to perfectly align with the strawberries on X,Y-axis.

\end{itemize}

%% file: results.tex
\section{Field experiments results}
\label{sec:results}
In order to validate the integrity of the system and verify its accuracy, field tests were carried out at two different sites. First, experiments were conducted at Berry Gardens strawberry poly-tunnels at the Riseholme campus of the University of Lincoln 2021. The second field test was carried out at a commercial strawberry growing glasshouse facility owned by Dyson Farming in Carrington, England in 2022. Figure \ref{fig:field_test} and \ref{fig:thorvald} show the system harvesting at Dyson glasshouse and Berry Gardens respectively. As can be seen at Berry Gardens the harvesting system was mounted on Thorvald mobile robot, and at Dyson glasshouse, it was mounted on a commercial harvesting trolley.

The Berry Gardens poly-tunnels is a research strawberry farming facility with two main strawberry varieties Driscoll Zara and Driscoll Katrina. The variety Zara has a longer calyx as compared to the Katrina variety which makes it more complicated for robotic harvesting. The mean diameter for the Katrina variety is 1.75 mm with a standard deviation of 0.24 mm, and for the Zara is 1.76 mm with a standard deviation of 0.25 mm \cite{vishnu2022peduncle}. The strawberry variety at Dyson glasshouse is a commercial variety that is grown and available widely.

\begin{figure}[tb!]
\begin{center}
  \includegraphics[width=1\linewidth]{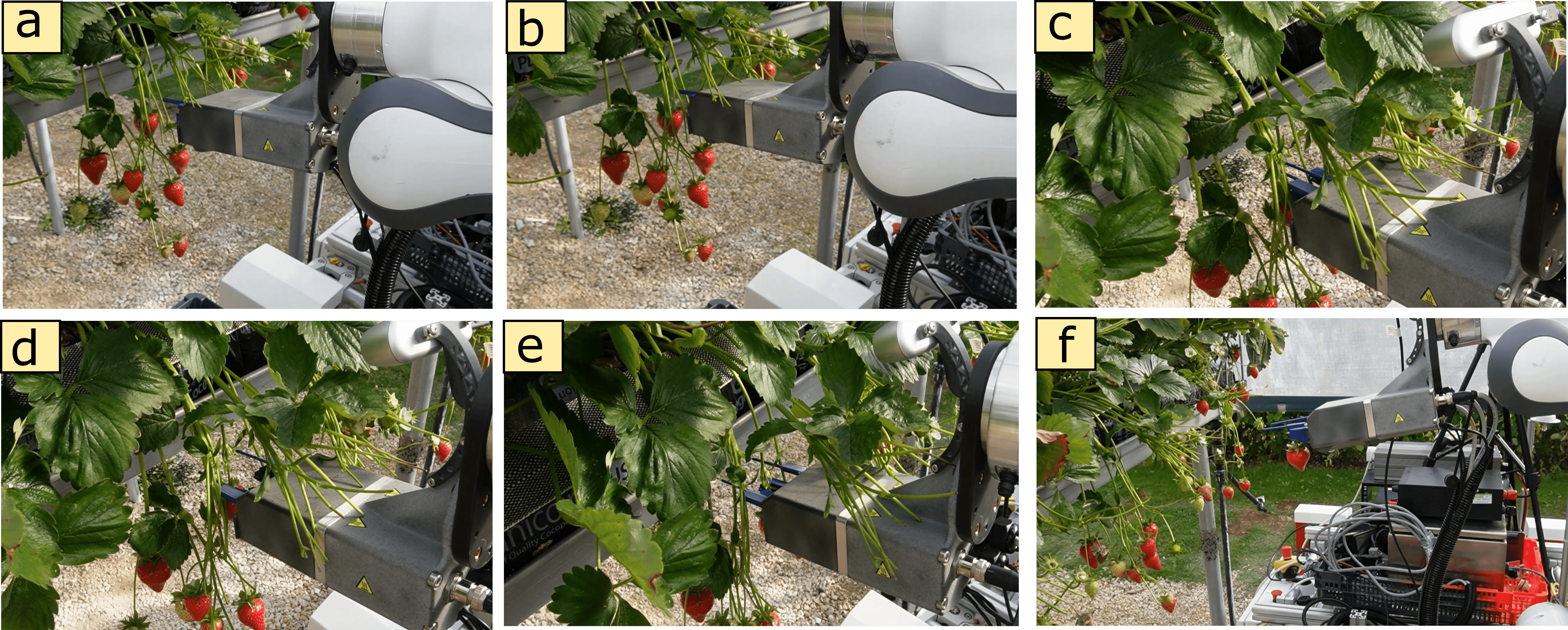}
\end{center}
  \caption{The harvesting sequences to remove a possible occlusion and pick fruit. The separator fingers penetrate in between the occluding fruits and by opening remove the occlusion and make way for gripping fingers to grasp the targeted fruit's stem and cut it.}
\label{fig:harvesting}
\end{figure}

Figure~\ref{fig:harvesting} demonstrates harvesting sequences including removing possible occlusion using separator fingers. The separator fingers penetrate in between the occluding fruits and by opening remove the occlusion and make way for gripping fingers to grasp the targeted fruit’s stem and cut it. In addition, the separator fingers can be used to remove detection occlusion as well to allow the perception system to detect all possible fruits.    

\subsection{Segmentation, Key-points and Pluckable Detection Results }
\label{sec:seg_kp_results}

Table \ref{table:segmentation} summarises the results for segmentation and  key-points detection of strawberries for both the datasets with Detectron-2 \cite{wu2019detectron2}. Although, our results demonstrate different backbones used in our experiments can produce consistent results across the dataset, ResNeXt based model performs better than ResNet-50 based model. The first two columns of~Table \ref{table:segmentation} show segmentation Average Precision (AP) values for pluckable and "unpluckable" berries separately. The sub-columns show AP for Intersection over Union (IoU), and thresholds of 0.5, 0.7, and 0.9. The standard practice is to consider IoU threshold 0.5 \cite{he2017mask}, however, we also show up to IoU threshold 0.9. Using Dataset-2, our proposed models yield decent AP values for both pluckable and "unpluckable" strawberries at IoU threshold 0.5. However, as shown in Table \ref{table:segmentation}, the `"unpluckable"' berries in Dataset-2 significantly outnumber the `pluckable' berries. This results in better segmentation performance of `"unpluckable"' berries. The performance drops significantly for stricter thresholds 0.7 and 0.9. This dataset represents berries in very dense clusters and thus Dataset-2 is a very challenging dataset and has the potential to further advance the research in selective harvesting. On the other hand, Dataset-1 shows very reliable AP values for pluckable strawberries for both the backbones across IoU thresholds. With IOU threshold of 0.5, the Detectron-2  produces 93.32 (R50-FPN) and (X101-FPN) 94.19 AP values, while with a very strict IoU threshold of 0.9 the Detectron-2 provides AP of 83.55, and 88.70 AP with R50-FPN and X101-FPN, respectively. This shows that for selective harvesting the dataset can be reliably used. For Dataset-1, the performance of our models on `"unpluckable"' berries is comparatively less reliable as there are fewer samples of `"unpluckable"' berries in this dataset. However, from a selective harvesting perspective instance segmentation of `pluckable' berries is more essential.

\begin{figure}[t!]
\begin{center}
  \includegraphics[width=0.7\linewidth]{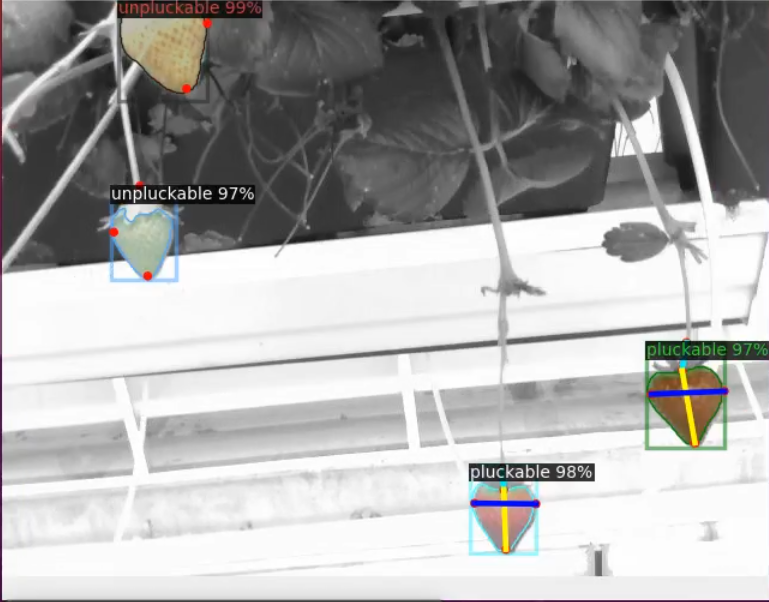}
\end{center}
  \caption{We introduce two novel datasets targeted toward the robotic selective harvesting of strawberries. The datasets provide instance segmentation, "pluckability", key-points, and weight information about the strawberries.}
  \vspace{0mm}
\label{fig:detection}
\end{figure}

The results of the key-points detection expressed in terms of AP at different IoU thresholds are similar to segmentation. At each IoU threshold, we take the average results from 0.5, 0.3, and 0.1 OKS. OKS \cite{wu2019detectron2} is the standard performance metric used by Detectron-2 \cite{wu2019detectron2} and MSCOCO \cite{lin2015microsoft} for key-point detection. While the OKS threshold normally used is 0.5, 0.1 is a stricter threshold. The experimental results show that similarly to segmentation, the results are consistent across the two backbones although X101-FPN performs slightly better. Also, the key-points detection for `pluckable' berries is much better than `"unpluckable"' berries for Dataset-1. The results for Dataset-2 obtained comparing X101-FPN and X101 networks, provide a good baseline for future research. Figure~\ref{fig:detection} shows an example of creating a fruit bounding box, key-points detection, and predicting pluckable and unpluckable fruits. 

\begin{table*}[!tb]
\begin{center}
\small\addtolength{\tabcolsep}{-2.5pt}
\caption{Segmentation and key-points detection results.The sub-columns show AP for Intersection over Union(IoU), and thresholds of 0.5, 0.7, and 0.9.}
\label{table:segmentation}
\resizebox{\textwidth}{!}{%
\begin{tabular}{lccccccccccccc}
\hline
\multirow{2}{*}{Dataset} & \multirow{2}{*}{Backbone} & \multicolumn{3}{c}{Segn Pluckable} & \multicolumn{3}{c}{Segn "unpluckable"} & \multicolumn{3}{c}{Key-points Pluckable} & \multicolumn{3}{c}{Key-points "unpluckable"}\\
& & 0.5 & 0.7 & 0.9 & 0.5 & 0.7 & 0.9 & 0.5 & 0.7 & 0.9 & 0.5 & 0.7 & 0.9\\
\hline
\multirow{2}{*}{Dataset-1} & R50-FPN & 93.32 & 90.97 & 83.55 & 59.46 & 53.61 & 42.91 & 91.27	 & 89.10 & 81.90 & 51.36 & 46.20 & 37.30\\
& X101 & 94.19 & 92.83 & 88.70 & 61.12 & 56.22 & 45.64 & 92.71 & 91.40 & 87.74 & 61.26 &	56.52 & 46.84\\
\hline
\multirow{2}{*}{Dataset-2} & X101-FPN & 71.12 &	64.70 &	43.24 &	76.83 &	74.52 &	68.79 & 64.32 &	58.93 & 39.92 & 73.26 & 71.39 & 66.46\\
& X101 & 72.12 & 66.84 & 47.86 & 78.09 & 76.65 & 70.30 & 59.29 & 54.40 & 42.12 & 74.67 &	71.45 & 65.30\\
\hline
\end{tabular}%
}
\end{center}
\vspace{-2mm}
\end{table*}

\subsection{Gaussian Process Regression for picking point error estimation results}

\begin{figure}[t!]
\begin{center}
  \includegraphics[width=1\linewidth]{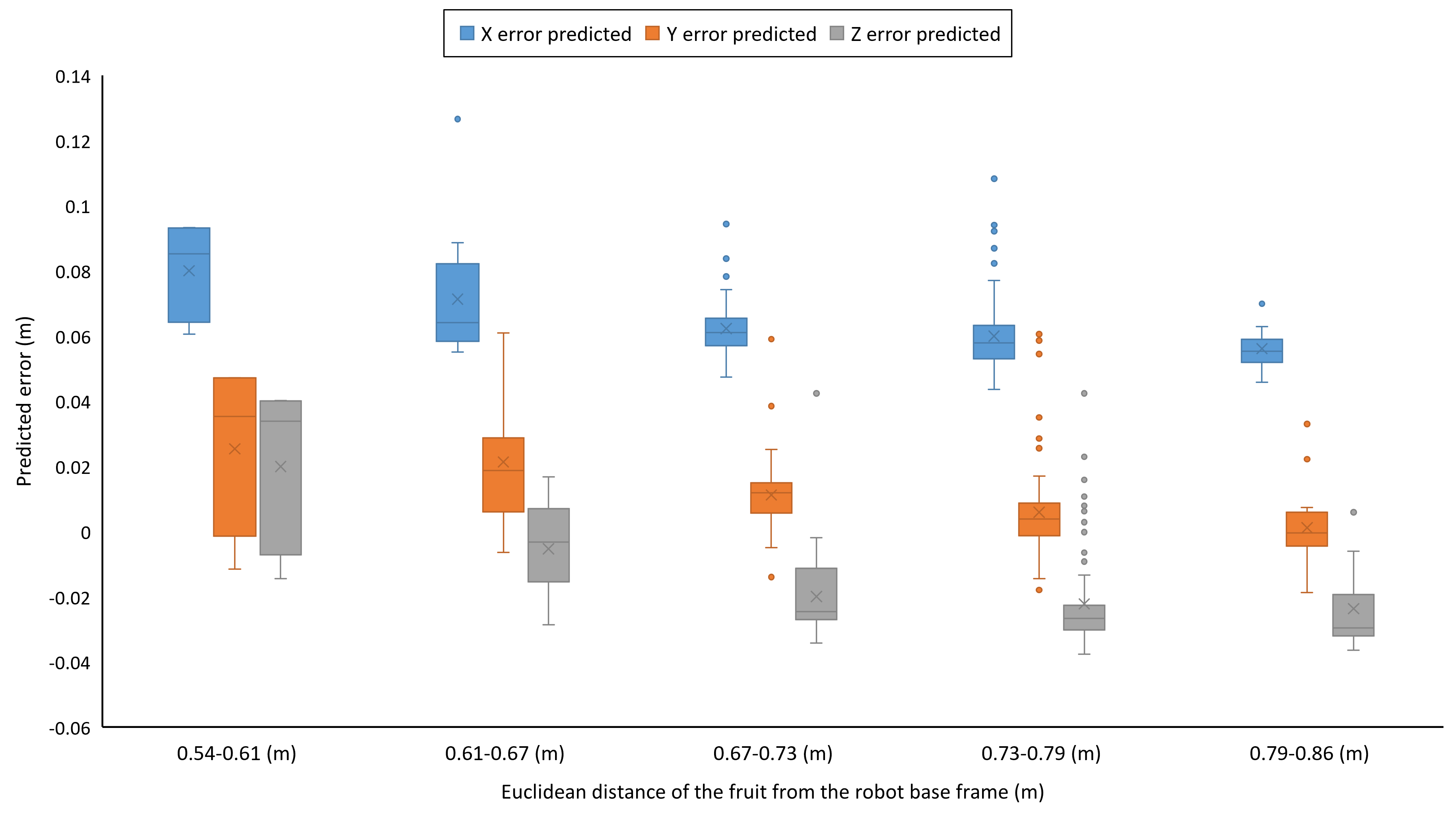}
\end{center}
  \caption{Predicted x, y, and z error of the fruit coordinate in the robot base frame using Gaussian Process Regression model.}
  \vspace{0mm}
\label{fig:error}
\end{figure}

During the field experiment, the results of picking point error estimation using the Gaussian Process Regression model were recorded which are shown in Figure~\ref{fig:error}. It presents the predicted x, y, and z error of the fruit coordinate in the robot base
frame corresponding to the euclidean distance of the fruit from the base of the robot. After predicting the errors using the model, the targeted fruit picking point coordinate is corrected and the new coordinate is sent to the system to re-plan the trajectory of the end-effector. The results show that the values of the errors are fluctuating around a specific mean value for each axis. The Mean of errors of x-axis is $ME_x=0.062 m$ with a standard deviation of $\sigma_x=0.012$, for y-axis is $ME_y=0.009 m$ with a standard deviation of $\sigma_y=0.014$, and for z-axis is $ME_z=-0.019 m$ with a standard deviation of $\sigma_z=0.016$.

\subsection{Harvesting analyses}

We perform a series of experiments where at each attempt we consider all the berries in a series of clusters. This typically means a target range from around 5 to 15 ripe fruits in a cluster among other yet to ripe fruit. Different trails in multiple harvesting sessions were carried out. The data collected from harvesting trails are presented in Table~\ref{table:performance}. For this experiment, first all fruit $N_a$ in the harvesting section including ripe and unripe fruits were counted. To determine the performance of the fruit ripeness detection of the system, the pluckable fruits $N_p$, also were counted based on human judgment and compared with the system's ripeness detection $N_d$. The ratio of pluckable fruit to all fruits including unripe fruits $(N_a/N_p)$ was 0.42. The result of calculating ripeness detection ratio $(N_d/N_p)$ shows 95\% accuracy of our ripeness detection model.

The trials show that the performance of the robot seemed to be influenced by the position of the fruit in the cluster, which varies significantly from one variety to another. The failure of the robot increases if the target fruit is occluded by too many fruit and/or leaves. However, our novel design of the end-effector, equipped with separator fingers, is able to unblock the most common occlusion.

The successfully harvested fruits were counted during the harvesting trails. A harvest attempt was considered a successful harvest where a ripe fruit was detected, griped, cut, and put in the punnet without damage or bruise. Where a fruit dropped midway, bruised or damaged with cutting or gripping, or harvested with a too long stem remaining on the fruit, considered an unsuccessful harvest. Therefore, the successful harvesting rate $S_r$ was calculated as $S_r=(N_s/N_p)\times100$ where $N_s$ is the total harvested fruits, and $N_p$ is the total pluckable fruit. The results presented in Table~\ref{table:performance} show $S_r=83\%$ for all trials. Another important parameter is the success rate of detected fruits which calculate as $SD_r=(N_d/N_p)\times100$ where $N_d$ is the total number of detected fruit by the system detection model, and $N_p$ is the total pluckable fruit. This parameter shows the performance of the designed end-effector, position error estimation model, and motion planning control system, regardless of whether fruit is detected or not. The results show $SD_r=87\%$ for all trials.

\begin{table}[H]
\begin{center}
\caption{Comparison of picking success rate after the first attempt with other strawberry harvesting systems. *Success of peduncle detection}
\label{table:picking_success}
\begin{tabular}{l|c}
\hline
Author & Success rate (\%)\\
\hline
\cite{dimeas2015design} & 65\\ 
\cite{hayashi2010evaluation} & 65*\\ 
\cite{feng2012new} & 71\\
\cite{ge2019fruit} & 74\\
This work $(S_r)$ & 83\\
This work $(SD_r)$ & 87\\
\hline
\end{tabular}
\end{center}
\end{table}

Table~\ref{table:picking_success} shows the performance of our system in comparison to existing approaches that have evaluated their systems. Among other methods only \cite{ge2019fruit} perform close in comparison to this work. It is not surprising that \cite{ge2019fruit} also relies on mask-RCNN for strawberry detection. Moreover, they also propose their own algorithm for refining the depth information obtained from the depth sensor. 

To analyze the unsuccessful harvests, five more parameters were defined and recorded during field experiments; total attempt $A_t$, cut command failure $F_c$, gripping/cutting failure $F_{gc}$, picking validation failure $F_v$, and position failure $F_p$. The total attempt is the sum of all robot attempts to harvest all detected fruits in a trail. Cut command failure is when the robot end-effector is successfully positioned at the picking point, but the system fails to detect the fruit and send a cutting command. The gripping/cutting failure parameter shows unsuccessful or partially cutting off the stem, and/or unsuccessful gripping leads to harvesting failure. Picking validation failure happens when the stem of the targeted fruit is gripped and cut successfully, but the system fails to validate picking and doesn't put the fruit in the punnet. The position failure parameter shows the failures due to inaccurate localizing of the picking point where the end-effector fails to grasp and/or cut the stem of the targeted fruit.

The results show in total 201 attempts were conducted by the robot to harvest all detected pluckable fruit which is 1.23 attempts per fruit. It can be seen that there are 66 failed attempts where 12\% is the result of detection failure, 29\% due to cut command failure, 14\% because of gripping/cutting failure, 9\% is the result of picking validation failure, and finally, 26\%  was the result of position inaccuracy.

\begin{figure}[t!]
\begin{center}
  \includegraphics[width=0.7\linewidth]{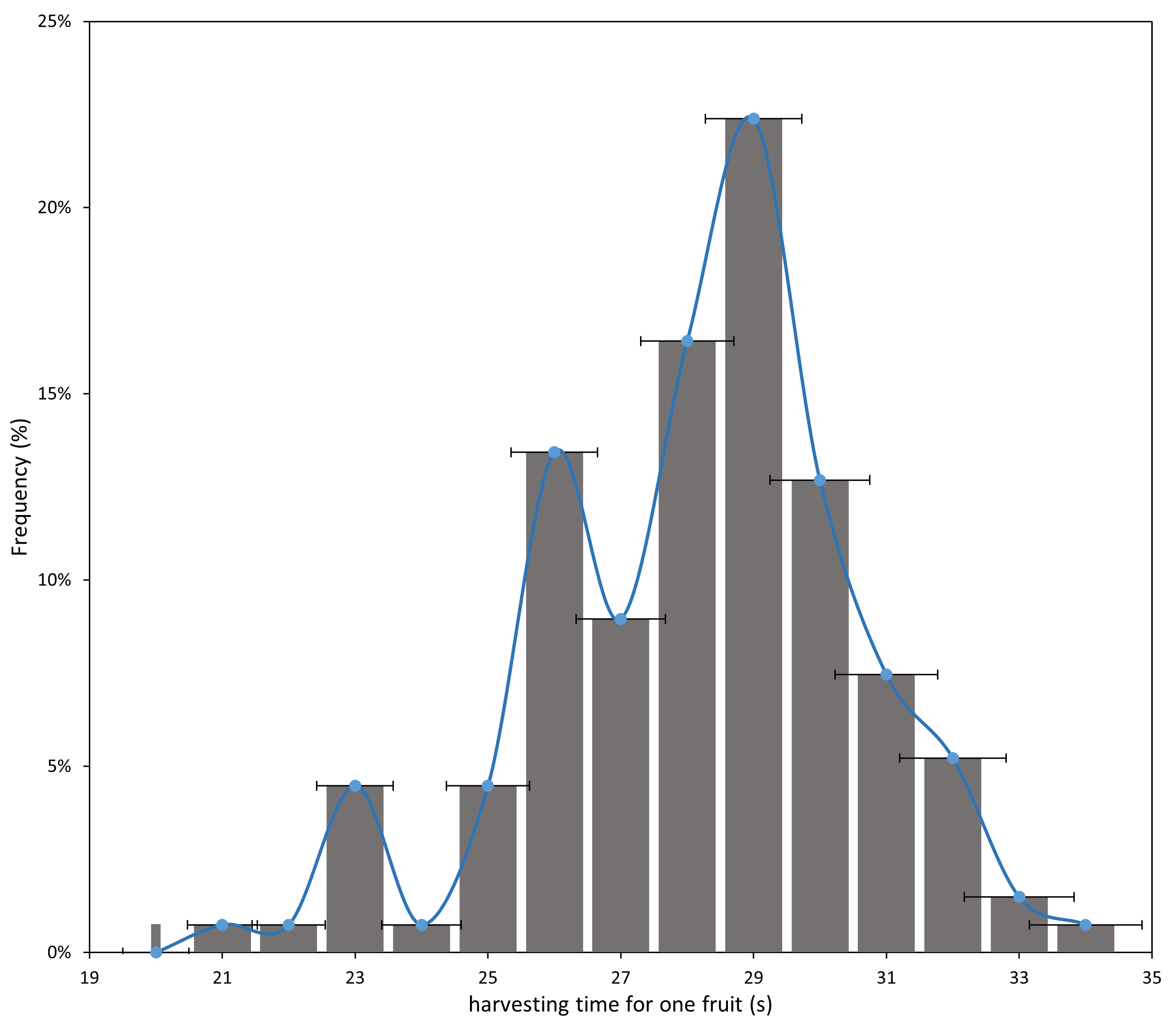}
\end{center}
  \caption{The distribution of time of picking one fruit. The time was measured from the beginning of capturing the image of the targeted fruit using the vision sensor to the end of placing the harvested fruit in the punnet.}
  \vspace{0mm}
\label{fig:time}
\end{figure}

The histogram of the time of picking one fruit successfully is presented in Figure~\ref{fig:time}. The time of harvest of one fruit here was measured from the capture of a fruit image to put that fruit in the punnet. The average execution time from capturing the image to placing the fruit was 28.2 s. In addition, the total time of a trial was measured as can be seen in Table~\ref{table:performance}. This time was measured from starting the robot at the beginning of a trail to the last harvest of that trail. The average time per successful harvest of this measurement is slightly higher than the average picking time. This is due to failed/delayed defections, failed attempts, etc.

\begin{table}[!ht]
\begin{center}
\caption{Results of field experiments.}

\label{table:performance}
\resizebox{\textwidth}{!}{%
\begin{tabular}{lllllllllll}
\hline
\textbf{Trail No.}  & \textbf{\begin{tabular}[c]{@{}l@{}}Total\\ fruit\end{tabular}} & \textbf{\begin{tabular}[c]{@{}l@{}}Pluckable\\    fruit\end{tabular}} & \textbf{\begin{tabular}[c]{@{}l@{}}Pluckable\\ not detected\end{tabular}} & \textbf{\begin{tabular}[c]{@{}l@{}}Cut comm.\\ Failure\end{tabular}} & \textbf{\begin{tabular}[c]{@{}l@{}}Grip/Cut\\ failure\end{tabular}} & \textbf{\begin{tabular}[c]{@{}l@{}}Picking\\ Valid. failure\end{tabular}} & \textbf{\begin{tabular}[c]{@{}l@{}}Position\\ failure\end{tabular}} & \textbf{\begin{tabular}[c]{@{}l@{}}Successful\\ harvest\end{tabular}} & \textbf{\begin{tabular}[c]{@{}l@{}}Total\\ attempt\end{tabular}} & \textbf{\begin{tabular}[c]{@{}l@{}}Total trail\\  time (s)\end{tabular}} \\ \hline
\multirow{2}{*}{1}  & \multirow{2}{*}{17}                                            & \multirow{2}{*}{7}                                                    & \multirow{2}{*}{1}                                                        & \multirow{2}{*}{2}                                                   & \multirow{2}{*}{0}                                                  & \multirow{2}{*}{0}                                                        & \multirow{2}{*}{1}                                                  & \multirow{2}{*}{6}                                                    & \multirow{2}{*}{10}                                              & \multirow{2}{*}{265}                                                     \\
                    &                                                                &                                                                       &                                                                           &                                                                      &                                                                     &                                                                           &                                                                     &                                                                       &                                                                  &                                                                          \\
\multirow{2}{*}{2}  & \multirow{2}{*}{15}                                            & \multirow{2}{*}{7}                                                    & \multirow{2}{*}{0}                                                        & \multirow{2}{*}{1}                                                   & \multirow{2}{*}{1}                                                  & \multirow{2}{*}{1}                                                        & \multirow{2}{*}{3}                                                  & \multirow{2}{*}{5}                                                    & \multirow{2}{*}{11}                                              & \multirow{2}{*}{355}                                                     \\
                    &                                                                &                                                                       &                                                                           &                                                                      &                                                                     &                                                                           &                                                                     &                                                                       &                                                                  &                                                                          \\
\multirow{2}{*}{3}  & \multirow{2}{*}{18}                                            & \multirow{2}{*}{8}                                                    & \multirow{2}{*}{1}                                                        & \multirow{2}{*}{0}                                                   & \multirow{2}{*}{0}                                                  & \multirow{2}{*}{0}                                                        & \multirow{2}{*}{0}                                                  & \multirow{2}{*}{7}                                                    & \multirow{2}{*}{7}                                               & \multirow{2}{*}{245}                                                     \\
                    &                                                                &                                                                       &                                                                           &                                                                      &                                                                     &                                                                           &                                                                     &                                                                       &                                                                  &                                                                          \\
\multirow{2}{*}{4}  & \multirow{2}{*}{24}                                            & \multirow{2}{*}{10}                                                   & \multirow{2}{*}{0}                                                        & \multirow{2}{*}{1}                                                   & \multirow{2}{*}{0}                                                  & \multirow{2}{*}{0}                                                        & \multirow{2}{*}{3}                                                  & \multirow{2}{*}{8}                                                    & \multirow{2}{*}{15}                                              & \multirow{2}{*}{441}                                                     \\
                    &                                                                &                                                                       &                                                                           &                                                                      &                                                                     &                                                                           &                                                                     &                                                                       &                                                                  &                                                                          \\
\multirow{2}{*}{5}  & \multirow{2}{*}{25}                                            & \multirow{2}{*}{12}                                                   & \multirow{2}{*}{1}                                                        & \multirow{2}{*}{2}                                                   & \multirow{2}{*}{1}                                                  & \multirow{2}{*}{1}                                                        & \multirow{2}{*}{1}                                                  & \multirow{2}{*}{11}                                                   & \multirow{2}{*}{15}                                              & \multirow{2}{*}{453}                                                     \\
                    &                                                                &                                                                       &                                                                           &                                                                      &                                                                     &                                                                           &                                                                     &                                                                       &                                                                  &                                                                          \\
\multirow{2}{*}{6}  & \multirow{2}{*}{14}                                            & \multirow{2}{*}{9}                                                    & \multirow{2}{*}{1}                                                        & \multirow{2}{*}{0}                                                   & \multirow{2}{*}{0}                                                  & \multirow{2}{*}{0}                                                        & \multirow{2}{*}{0}                                                  & \multirow{2}{*}{8}                                                    & \multirow{2}{*}{8}                                               & \multirow{2}{*}{231}                                                     \\
                    &                                                                &                                                                       &                                                                           &                                                                      &                                                                     &                                                                           &                                                                     &                                                                       &                                                                  &                                                                          \\
\multirow{2}{*}{7}  & \multirow{2}{*}{28}                                            & \multirow{2}{*}{12}                                                   & \multirow{2}{*}{2}                                                        & \multirow{2}{*}{1}                                                   & \multirow{2}{*}{0}                                                  & \multirow{2}{*}{2}                                                        & \multirow{2}{*}{0}                                                  & \multirow{2}{*}{9}                                                    & \multirow{2}{*}{15}                                              & \multirow{2}{*}{335}                                                     \\
                    &                                                                &                                                                       &                                                                           &                                                                      &                                                                     &                                                                           &                                                                     &                                                                       &                                                                  &                                                                          \\
\multirow{2}{*}{8}  & \multirow{2}{*}{8}                                             & \multirow{2}{*}{3}                                                    & \multirow{2}{*}{0}                                                        & \multirow{2}{*}{0}                                                   & \multirow{2}{*}{0}                                                  & \multirow{2}{*}{0}                                                        & \multirow{2}{*}{0}                                                  & \multirow{2}{*}{3}                                                    & \multirow{2}{*}{5}                                               & \multirow{2}{*}{78}                                                      \\
                    &                                                                &                                                                       &                                                                           &                                                                      &                                                                     &                                                                           &                                                                     &                                                                       &                                                                  &                                                                          \\
\multirow{2}{*}{9}  & \multirow{2}{*}{14}                                            & \multirow{2}{*}{6}                                                    & \multirow{2}{*}{0}                                                        & \multirow{2}{*}{2}                                                   & \multirow{2}{*}{0}                                                  & \multirow{2}{*}{0}                                                        & \multirow{2}{*}{0}                                                  & \multirow{2}{*}{5}                                                    & \multirow{2}{*}{10}                                               & \multirow{2}{*}{221}                                                     \\
                    &                                                                &                                                                       &                                                                           &                                                                      &                                                                     &                                                                           &                                                                     &                                                                       &                                                                  &                                                                          \\
\multirow{2}{*}{10} & \multirow{2}{*}{35}                                            & \multirow{2}{*}{18}                                                   & \multirow{2}{*}{0}                                                        & \multirow{2}{*}{0}                                                   & \multirow{2}{*}{0}                                                  & \multirow{2}{*}{1}                                                        & \multirow{2}{*}{0}                                                  & \multirow{2}{*}{16}                                                   & \multirow{2}{*}{17}                                              & \multirow{2}{*}{448}                                                     \\
                    &                                                                &                                                                       &                                                                           &                                                                      &                                                                     &                                                                           &                                                                     &                                                                       &                                                                  &                                                                          \\
\multirow{2}{*}{11} & \multirow{2}{*}{31}                                            & \multirow{2}{*}{9}                                                    & \multirow{2}{*}{0}                                                        & \multirow{2}{*}{4}                                                   & \multirow{2}{*}{1}                                                  & \multirow{2}{*}{0}                                                        & \multirow{2}{*}{1}                                                  & \multirow{2}{*}{8}                                                    & \multirow{2}{*}{14}                                              & \multirow{2}{*}{311}                                                     \\
                    &                                                                &                                                                       &                                                                           &                                                                      &                                                                     &                                                                           &                                                                     &                                                                       &                                                                  &                                                                          \\
\multirow{2}{*}{12} & \multirow{2}{*}{17}                                            & \multirow{2}{*}{10}                                                   & \multirow{2}{*}{0}                                                        & \multirow{2}{*}{2}                                                   & \multirow{2}{*}{0}                                                  & \multirow{2}{*}{0}                                                        & \multirow{2}{*}{1}                                                  & \multirow{2}{*}{8}                                                    & \multirow{2}{*}{11}                                              & \multirow{2}{*}{340}                                                     \\
                    &                                                                &                                                                       &                                                                           &                                                                      &                                                                     &                                                                           &                                                                     &                                                                       &                                                                  &                                                                          \\
\multirow{2}{*}{13} & \multirow{2}{*}{24}                                            & \multirow{2}{*}{11}                                                   & \multirow{2}{*}{0}                                                        & \multirow{2}{*}{3}                                                   & \multirow{2}{*}{1}                                                  & \multirow{2}{*}{0}                                                        & \multirow{2}{*}{2}                                                  & \multirow{2}{*}{8}                                                    & \multirow{2}{*}{12}                                              & \multirow{2}{*}{401}                                                     \\
                    &                                                                &                                                                       &                                                                           &                                                                      &                                                                     &                                                                           &                                                                     &                                                                       &                                                                  &                                                                          \\
\multirow{2}{*}{14} & \multirow{2}{*}{21}                                            & \multirow{2}{*}{6}                                                    & \multirow{2}{*}{1}                                                        & \multirow{2}{*}{3}                                                   & \multirow{2}{*}{1}                                                  & \multirow{2}{*}{0}                                                        & \multirow{2}{*}{2}                                                  & \multirow{2}{*}{4}                                                    & \multirow{2}{*}{10}                                              & \multirow{2}{*}{270}                                                     \\
                    &                                                                &                                                                       &                                                                           &                                                                      &                                                                     &                                                                           &                                                                     &                                                                       &                                                                  &                                                                          \\
\multirow{2}{*}{15} & \multirow{2}{*}{20}                                            & \multirow{2}{*}{9}                                                    & \multirow{2}{*}{0}                                                        & \multirow{2}{*}{1}                                                   & \multirow{2}{*}{1}                                                  & \multirow{2}{*}{1}                                                        & \multirow{2}{*}{2}                                                  & \multirow{2}{*}{7}                                                    & \multirow{2}{*}{10}                                              & \multirow{2}{*}{261}                                                     \\
                    &                                                                &                                                                       &                                                                           &                                                                      &                                                                     &                                                                           &                                                                     &                                                                       &                                                                  &                                                                          \\
\multirow{2}{*}{16} & \multirow{2}{*}{14}                                            & \multirow{2}{*}{7}                                                    & \multirow{2}{*}{1}                                                        & \multirow{2}{*}{2}                                                   & \multirow{2}{*}{0}                                                  & \multirow{2}{*}{0}                                                        & \multirow{2}{*}{0}                                                  & \multirow{2}{*}{5}                                                    & \multirow{2}{*}{8}                                               & \multirow{2}{*}{193}                                                     \\
                    &                                                                &                                                                       &                                                                           &                                                                      &                                                                     &                                                                           &                                                                     &                                                                       &                                                                  &                                                                          \\
\multirow{2}{*}{17} & \multirow{2}{*}{23}                                            & \multirow{2}{*}{8}                                                    & \multirow{2}{*}{0}                                                        & \multirow{2}{*}{1}                                                   & \multirow{2}{*}{2}                                                  & \multirow{2}{*}{0}                                                        & \multirow{2}{*}{0}                                                  & \multirow{2}{*}{8}                                                    & \multirow{2}{*}{11}                                              & \multirow{2}{*}{231}                                                     \\
                    &                                                                &                                                                       &                                                                           &                                                                      &                                                                     &                                                                           &                                                                     &                                                                       &                                                                  &                                                                          \\
\multirow{2}{*}{18} & \multirow{2}{*}{29}                                            & \multirow{2}{*}{11}                                                   & \multirow{2}{*}{0}                                                        & \multirow{2}{*}{1}                                                   & \multirow{2}{*}{1}                                                  & \multirow{2}{*}{0}                                                        & \multirow{2}{*}{1}                                                  & \multirow{2}{*}{9}                                                    & \multirow{2}{*}{12}                                              & \multirow{2}{*}{245}                                                     \\
                    &                                                                &                                                                       &                                                                           &                                                                      &                                                                     &                                                                           &                                                                     &                                                                       &                                                                  &                                                                          \\ \hline
\textbf{Total}      & \textbf{377}                                                   & \textbf{163}                                                          & \textbf{8}                                                                & \textbf{26}                                                          & \textbf{9}                                                          & \textbf{6}                                                                & \textbf{17}                                                         & \textbf{135}                                                          & \textbf{201}                                                     & \textbf{5324 (s)}                                                        \\ \hline
\end{tabular}%
}

\end{center}
\end{table}